\PassOptionsToPackage{unicode}{hyperref}
\PassOptionsToPackage{hyphens}{url}
\PassOptionsToPackage{dvipsnames,svgnames,x11names}{xcolor}
\documentclass[
  11pt,
]{article}
\usepackage{xcolor}
\usepackage[margin=1in]{geometry}
\usepackage{amsmath,amssymb}
\usepackage{iftex}
\ifPDFTeX
  \usepackage[T1]{fontenc}
  \usepackage[utf8]{inputenc}
  \usepackage{textcomp} % provide euro and other symbols
\else % if luatex or xetex
  \usepackage{unicode-math} % this also loads fontspec
  \defaultfontfeatures{Scale=MatchLowercase}
  \defaultfontfeatures[\rmfamily]{Ligatures=TeX,Scale=1}
\fi
\usepackage{lmodern}
\ifPDFTeX\else
\fi
\IfFileExists{upquote.sty}{\usepackage{upquote}}{}
\IfFileExists{microtype.sty}{% use microtype if available
  \usepackage[]{microtype}
  \UseMicrotypeSet[protrusion]{basicmath} % disable protrusion for tt fonts
}{}
\makeatletter
\@ifundefined{KOMAClassName}{% if non-KOMA class
  \IfFileExists{parskip.sty}{%
    \usepackage{parskip}
  }{% else
    \setlength{\parindent}{0pt}
    \setlength{\parskip}{6pt plus 2pt minus 1pt}}
}{% if KOMA class
  \KOMAoptions{parskip=half}}
\makeatother
\usepackage{color}
\usepackage{fancyvrb}

\DefineVerbatimEnvironment{Highlighting}{Verbatim}{commandchars=\\\{\}}
\newenvironment{Shaded}{}{}

\newcommand{\AttributeTok}[1]{\textcolor[rgb]{0.49,0.56,0.16}{#1}}

\newcommand{\BuiltInTok}[1]{\textcolor[rgb]{0.00,0.50,0.00}{#1}}

\newcommand{\CommentTok}[1]{\textcolor[rgb]{0.38,0.63,0.69}{\textit{#1}}}

\newcommand{\DataTypeTok}[1]{\textcolor[rgb]{0.56,0.13,0.00}{#1}}

\newcommand{\ExtensionTok}[1]{#1}

\newcommand{\NormalTok}[1]{#1}
\newcommand{\OperatorTok}[1]{\textcolor[rgb]{0.40,0.40,0.40}{#1}}

\newcommand{\StringTok}[1]{\textcolor[rgb]{0.25,0.44,0.63}{#1}}
\newcommand{\VariableTok}[1]{\textcolor[rgb]{0.10,0.09,0.49}{#1}}

\usepackage{longtable,booktabs,array}
\usepackage[labelformat=empty]{caption}
\let\oldlongtable\longtable
\let\endoldlongtable\endlongtable
\renewenvironment{longtable}{\footnotesize\oldlongtable}{\endoldlongtable\normalsize}
\usepackage{calc} % for calculating minipage widths
\usepackage{etoolbox}
\makeatletter
\patchcmd\longtable{\par}{\if@noskipsec\mbox{}\fi\par}{}{}
\makeatother
\IfFileExists{footnotehyper.sty}{\usepackage{footnotehyper}}{\usepackage{footnote}}
\makesavenoteenv{longtable}
\usepackage{graphicx}
\makeatletter
\newsavebox\pandoc@box
\newcommand*\pandocbounded[1]{% scales image to fit in text height/width
  \sbox\pandoc@box{#1}%
  \Gscale@div\@tempa{\textheight}{\dimexpr\ht\pandoc@box+\dp\pandoc@box\relax}%
  \Gscale@div\@tempb{\linewidth}{\wd\pandoc@box}%
  \ifdim\@tempb\p@<\@tempa\p@\let\@tempa\@tempb\fi% select the smaller of both
  \ifdim\@tempa\p@<\p@\scalebox{\@tempa}{\usebox\pandoc@box}%
  \else\usebox{\pandoc@box}%
  \fi%
}
\def\fps@figure{htbp}
\makeatother
\providecommand{\tightlist}{%
  \setlength{\itemsep}{0pt}\setlength{\parskip}{0pt}}
\usepackage[numbers,sort&compress]{natbib}
\usepackage{bookmark}
\IfFileExists{xurl.sty}{\usepackage{xurl}}{} % add URL line breaks if available
\hypersetup{
  pdftitle={The Calibration Floor: Format Repair Can Masquerade as Self-Correction at Small-to-Mid Scale},
  pdfauthor={Mingguang Chen --- DeepGrounding (corresponding: deepgroundingai@gmail.com); Bo Qu --- DeepGrounding; Licheng Wang --- AlphaAvatar},
  colorlinks=true,
  linkcolor={Maroon},
  filecolor={Maroon},
  citecolor={Blue},
  urlcolor={Blue},
  pdfcreator={LaTeX via pandoc}}

\title{The Calibration Floor: Format Repair Can Masquerade as
Self-Correction at Small-to-Mid Scale}
\author{Mingguang Chen\\
{\small DeepGrounding}\\
{\small \href{mailto:deepgroundingai@gmail.com}{deepgroundingai@gmail.com}}
\and
Bo Qu\\
{\small DeepGrounding}
\and
Licheng Wang\\
{\small AlphaAvatar}}
\date{July 2026}

\begin{document}
\maketitle

\begin{abstract}
Accuracy changes after language-model self-revision are usually
interpreted as changes in reasoning. We show this can fail at the
answer-extraction boundary, and test the failure causally rather than
only observationally. Across Qwen3.5 (0.8B-9B), Gemma-4-12B, and two
frontier models via API (Tencent Hy3, Nvidia Nemotron-3-Ultra-550B) in
29 primary cells plus a frontier arm, we decompose the always-revise
accuracy shift into a content margin (both answers parseable) and
format-recovery/loss margins (parseability changes). On 12 cells with
meaningful unparseable-answer rates, format effects exceed content
effects (Wilcoxon p=1.7e-3). To test this causally, we force
already-generated reasoning through grammar-constrained decoding so
every answer is parseable by construction: across 14 cells this closes a
median 71\% of the gap between the naive total effect and the
content-margin estimate, with two cells converging exactly and a
residual on the two largest-effect cells reported rather than dismissed.
A clustered model confirms floor-scale (0.8B/2B) models have far higher
odds of content-level change and harm than capable-scale models
(p\textless1e-7). Replicating a cited confidence-gating protocol
verbatim on Qwen3.5 does not reproduce its reported gain and shows the
same near-zero content margin. A frontier check on much larger models
shows format-dominance intensifying with scale: content margin is
exactly zero in all 5 cells despite total effects up to +0.275, though
this arm is lower-powered. The calibration-floor criterion on the
content margin reveals a squeeze: floor-scale cells have headroom but
insufficient signal, capable-scale cells have signal but little
headroom; only one cell is marginally viable, with negligible
sealed-holdout gain. Content is a minority share of what the field has
measured as self-correction. We release the instrument, code, and
derived results.
\end{abstract}

\section{1. Introduction}\label{introduction}

Intrinsic self-correction --- a model revising its own answer after a
self-generated critique, with no external label or verifier in the loop
--- is now a standard inference-time pattern
\citep{madaan2023selfrefine, shinn2023reflexion} and a building block of
training-time self-iteration
\citep{zelikman2022star, yuan2024selfrewarding}. Whether it helps or
hurts is one of the most cited negative results in the field: Huang et
al. \citep{huang2024cannotselfcorrect} showed that LLMs, asked to find
and fix their own reasoning errors \emph{without external feedback},
tend to degrade rather than improve. Kamoi et al.'s survey
\citep{kamoi2024when} catalogs when self-correction actually works and
identifies reliable error \emph{detection} as the recurring
prerequisite. Stav et al. \citep{stav2026when} add task sensitivity:
self-correction helps on some constraint-checkable tasks and fails on
open reasoning. A recent survey of recursive self-improvement
\citep{chen2026rsi} records this arc and leaves open exactly the
calibration question this paper answers: when does a model's own
probability mass align with correctness --- and, as it turns out, when
is an observed answer change even attributable to the model's
probability mass rather than to whether its free-form output happened to
be extractable.

We approach this literature with a calibration-floor apparatus: a
per-sample identity for the net gain of confidence-gated revision, and a
criterion for when a gate can possibly beat the ALWAYS-revise endpoint
that Huang et al.~measure (§3.2). That apparatus is the paper's formal
core. But applying it to real trajectories surfaces a confound the
identity itself does not model: \textbf{whether an answer can be
extracted from the model's free-form text is not the same question as
whether the model's answer is right}, and the rate at which extraction
succeeds can itself change between the initial answer and the revision
--- for reasons that have nothing to do with reasoning (token-budget
exhaustion, a review prompt that invites a longer re-derivation, a
placeholder the model echoes instead of filling in). When that
extraction-failure rate differs between \(a_0\) and \(a_f\), it
manufactures an accuracy delta that is scored identically to a genuine
repair or a genuine regression, silently contaminating every quantity in
§3.2's identity --- \(\mathrm{acc}_0\), the repair rate \(r\), the
damage rate \(d\) --- and every downstream conclusion drawn from it.

This is not a hypothetical concern. Running identical (model, task)
cells under two prompt regimes --- one truncation-prone, one repaired
(§4.9) --- \textbf{reverses the sign} of the apparent ALWAYS-revise
effect: under truncation-prone prompts the \emph{revision} frequently
fails to emit a parseable answer and is silently scored wrong
(Huang-consistent apparent harm); under repaired prompts the residual
failures concentrate in the \emph{initial} answer, so the revision's
``second chance'' at emitting a parseable string reads as repair
(apparent gain). Once the accuracy delta is decomposed into a content
margin (both answers parseable --- the only margin that can honestly be
attributed to reasoning) and two format margins, the content margin is
small and stable across the sign flip; essentially the entire swing is
format.

\textbf{What this paper is.} We keep the calibration-floor identity and
criterion (§3.2) as the formal engine, add the margin decomposition
(§3.3) that must be applied \emph{before} that engine is pointed at real
trajectories, and test three claims (§3.4): \textbf{C1} --- apparent
self-correction effects are dominated by extraction artifacts, in both
directions; \textbf{C2} --- once isolated, the content margin is
near-zero for capable models and real (and often harmful) only at floor
scale; \textbf{C3} --- confidence-gated selective revision at
\(\leq 12\)B has at most a marginal, single-task operating niche,
squeezed between insufficient signal at floor scale and absent headroom
at capable scale. We validate the identity/floor instrument on synthetic
sandboxes with known ground truth (§5.4), then test C1--C3 on 29 (model,
task) cells spanning three model families, using an
extraction-completeness admission gate, a forced-continuation probe, a
three-template paraphrase arm, a cross-family replication, and a
leave-one-out floor-prediction analysis.

\textbf{Contributions.} (1) An exact, additive decomposition of the
observed self-correction accuracy delta into a content margin and two
format margins, applicable to any always-revise trajectory with an
answer extractor. (2) A causal test of that decomposition ---
grammar-constrained re-extraction on already-generated reasoning ---
that moves the evidence for C1 beyond the observational limits of the
recover/loss categories, closing a median 71\% of the total-vs-content
gap on the cells where it matters most, with an honestly-reported
residual on two. (3) Empirical evidence, converging across six
independent checks (prompt-regime sign reversal, forced-continuation
probe, prompt paraphrase, the causal control, cross-family replication,
and a verbatim literature-protocol replication that does not reproduce
its source paper's gain), that apparent self-correction effects at
\(\leq 12\)B scale --- and, provisionally, at a
\textasciitilde55B-active frontier scale --- are dominated by the format
margins. (4) A scale contrast, now backed by a clustered (GEE) model
with well-separated odds ratios (\(16\)--\(21\times\), \(p<10^{-7}\))
plus checkpoint-level robustness checks (task-paired sign test, exact
checkpoint permutation, checkpoint-clustered bootstrap) that report
their own honest significance ceilings given only 4 distinct
checkpoints, rather than a small cell-level rank test alone, showing the
content margin is genuinely, and often harmfully, active at floor scale
(0.8B/2B) while being inert at capable scale (4B--12B) in the specific
checkpoints tested. (5) A re-derivation of the calibration-floor
criterion on the content margin specifically, including a nonparametric
correction that reverses two binormal-fit false positives, mapping every
cell onto a squeeze plane (Figure 5) whose viable-gating quadrant
contains exactly one marginal member. (6) A fully reproducible,
offline-recomputable trajectory library and instrument (extraction gate,
margin decomposition, causal control, statistical tests, atlas figures)
released alongside the paper.

\section{2. Related Work}\label{related-work}

\textbf{Self-refinement and its limits.} Self-Refine
\citep{madaan2023selfrefine} and Reflexion \citep{shinn2023reflexion}
established iterative self-critique; Huang et al.
\citep{huang2024cannotselfcorrect} showed that without external
feedback, average self-correction on reasoning tasks fails. Kamoi et al.
\citep{kamoi2024when} survey the conditions under which correction
succeeds; Tyen et al. \citep{tyen2024correct} decompose the bottleneck
into mistake \emph{finding} versus mistake \emph{fixing} and show that
LLMs can correct errors when the error location is given. Our identity
makes that decomposition algebraic: \(r\) is Tyen's fixability given a
triggered revision; TPR/FPR are the model's endogenous mistake-finding
rates under a confidence gate. Our margin decomposition (§3.3) adds a
layer beneath Tyen's: before asking whether a \emph{found} mistake gets
\emph{fixed}, one must ask whether the pre- and post-revision answers
are even comparably \emph{extractable} --- a question the fixability
literature does not raise because most benchmarks report accuracy
directly rather than accuracy conditioned on successful parsing. Stav et
al. \citep{stav2026when} attribute task-level variation to
verifiability; our floor-versus-capable contrast (§9.2) suggests some of
that variation is scale-confounded --- the same open task can show real
content-level instability at 0.8B and near-total inertia at 9B.

\textbf{Confidence-guided self-correction.} Li et al.
\citep{li2024confidence} identify over-criticism when models revise
high-confidence correct answers and propose IoE prompting. Kadavath et
al. \citep{kadavath2022know} establish that models can assess P(True)
for their own answers. These works \emph{demonstrate} that
confidence-aware gating can help on specific benchmarks. Our finding is
a caution on the measurement side of that literature: any accuracy delta
attributed to gated versus ungated revision should first be checked
against the margin decomposition, since a nonzero delta is consistent
with pure extraction noise even when the gating policy itself does
nothing.

\textbf{Selective prediction and calibration.} The risk--coverage
framework \citep{chow1957optimum, geifman2017selective} asks when a
classifier should abstain. We transfer the logic to self-correction:
abstention means ``keep \(a_0\)''; triggering means ``adopt the revised
answer.'' The floor generalizes ``AUROC \(> 0.5\) suffices'' to a
task-dependent slope \(\lambda\) set by \((\mathrm{acc}_0, r, d)\), and
the \(\delta\) correction blocks the degenerate corner where binormal
ROCs nominally cross \(\lambda\) at FPR \(\to 0\) with vanishing
\(\Delta\). §4.8/§9.3 show this criterion must be computed on the
content margin, or the format artifact reappears inside \(r\) and \(d\)
and silently degenerates AUROC*.

\textbf{Sequence likelihood and intrinsic evaluation floors.} Zenn and
Geiping \citep{zenn2026when} show sequence probability is a conservative
correctness proxy; SelfCheckGPT \citep{manakul2023selfcheckgpt} uses
sampling consistency for hallucination detection. A within-cell ranking
of the five intrinsic signals is out of scope here (the self-consistency
subsample fails the extraction gate; §6.1) and left as future work.

\textbf{Companion studies.} Concurrent preprints in the same series
manipulate evaluator error structure under fixed marginal accuracy
\citep{chen2026error} and study excess self-confirmation drift in closed
evaluation loops \citep{chen2026esc}. Those papers share the
frozen-trajectory harness and offline policy-evaluation engine; the
present paper's extraction-gate and margin-decomposition instrument is
directly reusable by both, since any pipeline that scores free-form
generations against an extracted answer is exposed to the same confound.

\textbf{Positioning.} Prior work shows confidence gating can help on
specific benchmarks and that self-correction fails on average. We are
not aware of prior work that isolates an extraction-artifact margin from
a genuine content margin in self-correction accuracy deltas, shows the
artifact can flip the sign of a headline finding on identical (model,
task) cells, and re-derives a calibration-floor criterion that is valid
once the artifact is removed.

\section{3. The Margin Decomposition and Calibration
Floor}\label{the-margin-decomposition-and-calibration-floor}

\subsection{3.1 Research questions}\label{research-questions}

Under strict no-external-feedback self-correction on small-to-mid
open-weight models (0.8B--12B class):

\begin{itemize}
\tightlist
\item
  \textbf{RQ1 (decomposition).} Can the observed self-correction
  accuracy delta be exactly decomposed into a content margin and format
  margins, and do apparent effects in the literature (and in naive runs
  of this harness) concentrate in the format margins?
\item
  \textbf{RQ2 (scale).} Once isolated, does the content margin behave
  differently across model scale --- inert at capable scale, real at
  floor scale?
\item
  \textbf{RQ3 (floor, corrected).} When the calibration-floor criterion
  is applied to the content margin only, where do the tested cells fall
  on the squeeze plane, and how large is the viable-gating region?
\end{itemize}

\subsection{3.2 The per-sample
identity}\label{the-per-sample-identity}

For each sample with initial correctness \(y_0 \in \{0,1\}\), final
correctness \(y_f\) after a fixed always-revise protocol, and a gate
that triggers revision when confidence falls below \(\tau\):

\begin{Shaded}
\begin{Highlighting}[]
\NormalTok{Delta(tau) = (1 {-} acc\_0) * TPR(tau) * r(tau)  {-}  acc\_0 * FPR(tau) * d(tau)}

\NormalTok{acc\_0 = P(y\_0 = 1)}
\NormalTok{TPR   = P(trigger | y\_0 = 0)           \# recall of wrong answers}
\NormalTok{FPR   = P(trigger | y\_0 = 1)           \# false trigger on correct answers}
\NormalTok{r     = P(y\_f = 1 | trigger, y\_0 = 0)  \# repair rate}
\NormalTok{d     = P(y\_f = 0 | trigger, y\_0 = 1)  \# damage rate}
\end{Highlighting}
\end{Shaded}

All four components are measured offline from one always-revise
trajectory per (model, task). Answer-space geometry enters through
\(d\): open-ended tasks have \(d\) near 1; \(K\)-way MCQ has
\(d \lesssim (K-1)/K\) under random wrong flips. Profitability at
\(\tau\) requires \(\mathrm{TPR}/\mathrm{FPR} > \lambda(\tau)\) with
\(\lambda = \mathrm{acc}_0\cdot d / ((1-\mathrm{acc}_0)\cdot r)\) when
\(r\) and \(d\) are approximately constant in \(\tau\) (tested; if not,
the exact \(\tau\)-dependent form still applies).

\textbf{ALWAYS endpoint.} At (FPR, TPR) = (1, 1),
\(\Delta_{\mathrm{ALWAYS}} = (1-\mathrm{acc}_0)\cdot r - \mathrm{acc}_0\cdot d\),
profitable iff \(\lambda < 1\). Huang et al.'s negative result is
consistent with open reasoning cells where \(\lambda > 1\) regardless of
calibration; calibration determines whether \emph{other} ROC points
rescue net gain.

\textbf{\(\delta\)-corrected floor.} For binormal ROCs, any AUROC
\(> 0.5\) nominally crosses \(\lambda\) at FPR \(\to 0\), but
\(\Delta \to 0\) there. We therefore define profitability as
\(\max_\tau \Delta(\tau) \geq \delta\) with \(\delta = 0.01\) fixed in
advance, and solve for the minimum AUROC* (the \textbf{calibration
floor}) numerically from \((\mathrm{acc}_0, r, d)\).

\subsection{3.3 The format/content margin
decomposition}\label{the-formatcontent-margin-decomposition}

Every sample's \((a_0\_\mathrm{answer}, a_f\_\mathrm{answer})\) pair,
after running the task's answer extractor, falls into exactly one of
four categories:

\begin{Shaded}
\begin{Highlighting}[]
\NormalTok{bothok   a0 parseable, af parseable    {-}\textgreater{} CONTENT margin: both{-}parseable answer{-}change margin}
\NormalTok{recover  a0 unparseable, af parseable  {-}\textgreater{} FORMAT{-}RECOVER: extraction got a second chance}
\NormalTok{loss     a0 parseable, af unparseable  {-}\textgreater{} FORMAT{-}LOSS: extraction lost a working answer}
\NormalTok{dead     both unparseable              {-}\textgreater{} scored wrong under both; contributes 0 to Delta\_total}
\end{Highlighting}
\end{Shaded}

Scoring is unconditional (an unparseable answer is graded wrong, as any
accuracy pipeline would), so the following decomposition is exact and
additive over the \emph{entire} sample, not just the parseable subset:

\begin{Shaded}
\begin{Highlighting}[]
\NormalTok{Delta\_total          = acc\_f {-} acc\_0                          (over all n samples)}
\NormalTok{Delta\_content         = P(bothok, y0=0, yf=1) {-} P(bothok, y0=1, yf=0)}
\NormalTok{Delta\_format\_recover  = P(recover, yf=1)         \# a0 was wrong{-}by{-}scoring, af supplies an answer}
\NormalTok{Delta\_format\_loss     = {-}P(loss,   y0=1)         \# a0 was right, af fails to supply an answer}

\NormalTok{Delta\_total = Delta\_content + Delta\_format\_recover + Delta\_format\_loss   \# exact identity}
\end{Highlighting}
\end{Shaded}

Only \(\Delta_{\mathrm{content}}\) can be attributed to a change in the
model's \emph{answer}; the format margins are attributable to whether
the model's free-form text happened to contain something the extractor
could read off, which is sensitive to token budget, prompt wording, and
incidental truncation (§4.9, §9.1). Within the \texttt{bothok} subset,
the identity of §3.2 applies unchanged, restricted to that subset:
\(r\), \(d\), TPR, FPR, and the floor criterion are all well-defined and
interpretable as genuine repair/damage rates only when computed there
(§4.8).

\textbf{A naming caveat.} We call \(\Delta_{\mathrm{content}}\) the
``content margin'' throughout for brevity, but it is more precisely the
\textbf{both-parseable answer-change margin}: \texttt{bothok} membership
is determined jointly by \(a_0\) and \(a_f\), i.e.~by an outcome of the
revision itself, so it is a post-treatment-selected subgroup rather than
a fixed, pre-specified population. The decomposition
\(\Delta_{\mathrm{total}} = \Delta_{\mathrm{content}} + \Delta_{\mathrm{format\text{-}recover}} + \Delta_{\mathrm{format\text{-}loss}}\)
is exact algebra over the full sample regardless of this selection ---
it is not a causal estimate and does not depend on \texttt{bothok} being
an unbiased subgroup. What is not exact is the further step of reading
\(\Delta_{\mathrm{content}}\) as ``the reasoning effect'': because the
subgroup is selected on an outcome, its own within-group repair/damage
rates (\(r\), \(d\) above) can differ from what an unselected population
would show, and the causal control of §4.12 addresses this gap only in
the narrower sense defined there, not by certifying \texttt{bothok} as
selection-free.

\textbf{Imputation bounds.} Because the ground truth of an unparseable
answer is unknown, \(\Delta_{\mathrm{content}}\) as measured is the
identity's value under the convention ``unparseable counts as wrong on
both sides of the comparison.'' Two bounds on the true content effect
follow immediately: a lower bound assuming every recovered answer was
already correct pre-extraction-failure (no real flip),
\(\Delta_{\mathrm{content}} - P(\mathrm{loss}, y_0=1)\), and an upper
bound assuming every recovered/lost row was a genuine flip,
\(\Delta_{\mathrm{content}} + P(\mathrm{recover}, y_f=1)\). The
forced-continuation probe (§4.10) resolves the imputation question
empirically on the cells where it matters most.

\subsection{3.4 Claims and falsification
conditions}\label{claims-and-falsification-conditions}

\textbf{C1 (format artifacts are the primary, but not sole, driver of
apparent effects, bidirectionally).} On cells with an active extraction
channel (\(\geq 5\%\) of initial answers unparseable),
\(|\Delta_{\mathrm{format\text{-}recover}} + \Delta_{\mathrm{format\text{-}loss}}|\)
exceeds \(|\Delta_{\mathrm{content}}|\), and this holds independent of
the sign of \(\Delta_{\mathrm{total}}\) --- the same mechanism can
manufacture an apparent gain or an apparent harm depending on which side
of the \(a_0 \to a_f\) transition the extractor happens to fail on.
Because the observational recover/loss categories cannot by themselves
rule out a genuine content change riding along with a parseability
change, a causal control (§4.12) freezes the already-generated reasoning
text and forces guaranteed-parseable re-extraction from it --- a test of
how much the total effect shrinks once structured output is imposed, not
a full identification of whether the model's free-form output already
carried the same answer (the forced re-extraction is itself a minimal
new elicitation): it should close most, but need not close all, of the
gap between \(\Delta_{\mathrm{total}}\) and
\(\Delta_{\mathrm{content}}\). \emph{Falsified if} the content margin
dominates on active-channel cells, if prompt changes that alter
extraction quality leave \(\Delta_{\mathrm{total}}\) unchanged, or if
the causal control's closure is small or inconsistent in sign.

\textbf{C2 (content inertia at capable scale, real flips at floor
scale).} Restricted to the content margin, capable models (4B--12B) show
\(|\Delta_{\mathrm{content}}| \leq 0.03\) and content-level change rate
\(\leq 0.05\) on non-ARC answer-level tasks; floor-scale models
(0.8B/2B) show materially higher change rate and a nonzero, often
net-harmful content effect on the same task families. \emph{Falsified
if} capable-scale cells show large or systematically positive content
effects, or if floor-scale cells are equally inert.

\textbf{C3 (squeeze: at most a marginal gating niche at \(\leq 12\)B).}
The calibration-floor criterion, computed on the content margin,
predicts that floor-scale cells fail the floor (best-signal AUROC below
the \(\delta\)-corrected threshold) despite having real flip headroom,
while capable-scale cells have near-zero content headroom (oracle \(-\)
\(\max(\mathrm{NEVER}, \mathrm{ALWAYS})\), content margin) regardless of
signal quality --- so the region of the (headroom, signal-surplus) plane
where selective gating could pay is at most marginally populated.
\emph{Falsified if} multiple cells, or any cell with substantial
headroom, exhibit both floor-passing signal and exploitable headroom.

Out of scope by design: within-cell signal ranking (requires the
self-consistency signal-D subsample, which fails the extraction
admission gate; §6.1), quantization robustness, and \(\tau\)-quantile
transfer across tasks. The explicit-vs-implicit dominance-gap machinery
is retained in the instrument (§5.4, Check 5) but not claimed as a
tested hypothesis: the pre-fix MCQ trajectories that motivated it are
extraction-contaminated (§4.9), and the post-fix capable-scale cells
have too few content-level flips for the implicit operating point to be
informative.

\section{4. Method}\label{method}

\subsection{4.1 Frozen-trajectory offline
evaluation}\label{frozen-trajectory-offline-evaluation}

Gating decides only whether to keep the initial answer \(a_0\) or adopt
the revision \(a_f\); it does not alter revision content. Therefore:

\begin{Shaded}
\begin{Highlighting}[]
\NormalTok{Per (model, task, prompt template): ONE always{-}revise run}
\NormalTok{  a\_0 {-}\textgreater{} review {-}\textgreater{} a\_1 {-}\textgreater{} review {-}\textgreater{} a\_2   (T = 2; early stop if answer unchanged)}

\NormalTok{Record per node: text, y, signals A/A\textquotesingle{}/B/C, changed flag}
\NormalTok{Offline (zero extra generation):}
\NormalTok{  NEVER, ALWAYS, all (signal, tau), per{-}sample oracle,}
\NormalTok{  identity components, floor criterion,}
\NormalTok{  margin decomposition (content / format{-}recover / format{-}loss)}
\end{Highlighting}
\end{Shaded}

Signal C (verbalized confidence) is collected on an \textbf{independent
probe branch} so self-assessment prompts do not contaminate the revision
context.

\subsection{4.2 Admission gates}\label{admission-gates}

\begin{itemize}
\tightlist
\item
  \textbf{Floor gate:} \(\mathrm{acc}_0 \in [0.20, 0.85]\) and
  \(\geq 60\) wrong / \(\geq 60\) correct on dev (else adjust difficulty
  or mark as floor cell).
\item
  \textbf{Rigidity gate:} implicit change rate
  \(P(\mathrm{changed}) \in [0.03, 0.97]\); else escalate to a stronger
  review template; still rigid \(\Rightarrow\) record as degenerate.
\item
  \textbf{Signal gate:} verbalized C with zero variance \(\Rightarrow\)
  mark degenerate, report degeneration rate.
\end{itemize}

A fourth gate --- extraction completeness --- was added post hoc after
the first grid audit and is described in §4.9. Because this rule was
informed by observed failures, all analyses depending on the
admitted-cell set inherit that limitation.

\subsection{4.3 Models}\label{models}

Local inference via MLX (\texttt{mlx-lm}) for the Qwen3.5 family, 4-bit
throughout; the Gemma-4 family check runs via a local \texttt{ollama}
server (\texttt{think:false}, greedy, native logprobs).

{\def\LTcaptype{none} % do not increment counter
\begin{longtable}[]{@{}
  >{\raggedright\arraybackslash}p{(\linewidth - 4\tabcolsep) * \real{0.3333}}
  >{\raggedright\arraybackslash}p{(\linewidth - 4\tabcolsep) * \real{0.3333}}
  >{\raggedright\arraybackslash}p{(\linewidth - 4\tabcolsep) * \real{0.3333}}@{}}
\toprule\noalign{}
\begin{minipage}[b]{\linewidth}\raggedright
Role
\end{minipage} & \begin{minipage}[b]{\linewidth}\raggedright
Model
\end{minipage} & \begin{minipage}[b]{\linewidth}\raggedright
Use
\end{minipage} \\
\midrule\noalign{}
\endhead
\bottomrule\noalign{}
\endlastfoot
Primary & Qwen3.5-4B (non-thinking) & Full grid, all signals \\
Scale & Qwen3.5-9B (non-thinking) & Matched primary tasks \\
Floor & Qwen3.5-0.8B / Qwen3.5-2B & Content-margin scale contrast
(C2/C3) \\
Family check & Gemma-4-12B (via \texttt{ollama}, \texttt{think:false}) &
GSM8K / MMLU / MATH, cross-family C1/C2 replication \\
Frontier check & Tencent Hy3, Nvidia Nemotron-3-Ultra-550B (via
OpenRouter API, free tier) & GSM8K / MMLU / MATH,
scale-external-validity arm (§4.15, §9.8) \\
\end{longtable}
}

The original design anchored Qwen2.5-Instruct 0.5B/3B/7B. Before the
margin-decomposition pivot, we substituted Qwen3.5 (0.8B/2B/4B/9B,
Apache 2.0, \texttt{mlx-community} 4-bit builds), preserving the
intended size classes and disabling thinking mode
\citep{qwen2026qwen35omni, yang2025qwen3}. The later C1--C3 analysis
change is documented separately in §6.3 and must not be conflated with
this checkpoint substitution. Gemma-4-12B was added after the pivot
specifically to test whether the C1/C2 pattern was Qwen-specific. The
two frontier models were added later still, specifically to test whether
the squeeze (C3) opens up at a scale far beyond anything locally
hostable on the study's Apple Silicon hardware (24GB unified memory,
which cannot fit a 70B-class model even at 4-bit); §4.15 documents the
access route and its consequences for data completeness.

\subsection{4.4 Tasks}\label{tasks}

{\def\LTcaptype{none} % do not increment counter
\begin{longtable}[]{@{}
  >{\raggedright\arraybackslash}p{(\linewidth - 8\tabcolsep) * \real{0.2000}}
  >{\raggedright\arraybackslash}p{(\linewidth - 8\tabcolsep) * \real{0.2000}}
  >{\raggedright\arraybackslash}p{(\linewidth - 8\tabcolsep) * \real{0.2000}}
  >{\raggedright\arraybackslash}p{(\linewidth - 8\tabcolsep) * \real{0.2000}}
  >{\raggedright\arraybackslash}p{(\linewidth - 8\tabcolsep) * \real{0.2000}}@{}}
\toprule\noalign{}
\begin{minipage}[b]{\linewidth}\raggedright
Family
\end{minipage} & \begin{minipage}[b]{\linewidth}\raggedright
Pool
\end{minipage} & \begin{minipage}[b]{\linewidth}\raggedright
Scoring
\end{minipage} & \begin{minipage}[b]{\linewidth}\raggedright
n / model
\end{minipage} & \begin{minipage}[b]{\linewidth}\raggedright
Geometry
\end{minipage} \\
\midrule\noalign{}
\endhead
\bottomrule\noalign{}
\endlastfoot
Math (open) & GSM8K \citep{cobbe2021training}; MATH L1--3
\citep{hendrycks2021math} & exact / approximate normalized match & 400
each (300 at floor scale) & \(d\) near 1, high \(\lambda\) \\
Code & HumanEval \citep{chen2021humaneval} + MBPP-sanitized
\citep{austin2021mbpp} & unit tests & 400 & \(d\) near 1, \(r\) often
higher \\
MCQ & MMLU (4 subjects) \citep{hendrycks2021mmlu}; ARC-Challenge
\citep{clark2018arc} & option match & 400 each (300 at floor scale) &
low \(d\), low \(\lambda\) \\
Short answer & TriviaQA short \citep{joshi2017triviaqa} & normalized
match & 400 & intermediate \\
Robustness & CommonsenseQA \citep{talmor2019commonsenseqa};
TruthfulQA-MC1 \citep{lin2022truthfulqa} & option match & 200 each &
secondary MCQ cells \\
\end{longtable}
}

50/50 dev/holdout split per task, seed 2026, frozen before generation.
Comparisons are within-model across strategies; benchmark contamination
is a stated limitation.

\subsection{4.5 Self-correction
protocol}\label{self-correction-protocol}

\begin{Shaded}
\begin{Highlighting}[]
\NormalTok{Round 0: CoT prompt {-}\textgreater{} a\_0; record token logprobs (signals A, B)}
\NormalTok{Round n in \{1,2\}: feed full prior answer + review template T1}
\NormalTok{  "Review your reasoning above; fix errors if needed, else keep the answer."}
\NormalTok{  (no correctness information)}
\NormalTok{Early stop if answer region unchanged}
\NormalTok{Probe branch (not in revision context):}
\NormalTok{  C: verbalized 0{-}100 confidence}
\NormalTok{  A\textquotesingle{}: teacher{-}force P(True) on "Is the above answer correct? Yes/No"}
\end{Highlighting}
\end{Shaded}

Main decoding: greedy (\(T=0\)). \texttt{code/cf\_core.py} implements
all offline metrics.

\subsection{4.6 Intrinsic confidence
signals}\label{intrinsic-confidence-signals}

All target the round-0 answer.

\begin{Shaded}
\begin{Highlighting}[]
\NormalTok{A   token{-}logprob aggregate on answer span (geometric mean; MCQ: softmax over options)}
\NormalTok{A\textquotesingle{}  P(True): renormalized prob of "Yes" in Kadavath{-}style probe (one forward pass)}
\NormalTok{B   length{-}normalized sequence log{-}likelihood over full CoT (Zenn floor candidate)}
\NormalTok{C   verbalized 0{-}100 integer / 100}
\end{Highlighting}
\end{Shaded}

Signals enter gating by \textbf{rank} within cell (monotone invariance).
Per-cell ``best signal'' means highest wrongness-AUROC on the dev
split's content margin. This best-of-five choice was not used by the
sealed primary policy, which fixed \(p_{\mathrm{norm}}\); it is an
exploratory upper-bound analysis and is susceptible to dev-set selection
optimism.

\subsection{4.7 Gating strategies and two
estimands}\label{gating-strategies-and-two-estimands}

\begin{Shaded}
\begin{Highlighting}[]
\NormalTok{gate(x; S, tau): adopt revision iff conf\_S(x) \textless{} tau}

\NormalTok{Primary sealed policy:}
\NormalTok{  signal = p\_norm fixed before decryption}
\NormalTok{  tau*: dev grid over confidence quantiles (step 0.05), maximize raw{-}margin Delta\_dev}
\NormalTok{  evaluate once on holdout (physically isolated via seal/select{-}tau/decrypt{-}eval)}

\NormalTok{Exploratory floor/squeeze estimand:}
\NormalTok{  choose the highest{-}AUROC signal on the dev content margin}
\NormalTok{  evaluate content{-}margin max\_tau Delta and AUROC{-}vs{-}floor geometry}

\NormalTok{Baselines: NEVER, ALWAYS, random{-}gate (matched trigger rate), per{-}sample oracle}
\end{Highlighting}
\end{Shaded}

The sealed policy and the exploratory squeeze analysis answer different
questions and are not interchangeable. The former estimates deployable
gain for a prespecified signal on the ordinary benchmark score; the
latter asks whether any measured intrinsic signal could, in principle,
clear the calibration floor after removing extraction artifacts.

\subsection{4.8 Floor criterion, computed on the content
margin}\label{floor-criterion-computed-on-the-content-margin}

\begin{Shaded}
\begin{Highlighting}[]
\NormalTok{lambda = acc\_0 * d / ((1 {-} acc\_0) * r)                 \# all computed on the bothok subset}
\NormalTok{AUROC* = min AUROC (binormal fit) s.t. max\_tau Delta(tau) \textgreater{}= delta   \# delta = 0.01}
\NormalTok{floor\_passes iff AUROC\_best\_dev(content) \textgreater{} AUROC*(content)}
\end{Highlighting}
\end{Shaded}

Computing \(\lambda\)/AUROC* on the raw (non-decomposed) trajectory
instead of the content margin lets format-recovery inflate the apparent
repair rate \(r\), which can push AUROC* to a near-zero,
trivially-passable value --- we observed this directly on two MCQ floor
cells (AUROC* \textasciitilde0.002 on the raw margin, versus a
well-defined non-degenerate threshold on the content margin). All floor
figures in §9.3 are content-margin figures. Where AUROC* is genuinely
undefined because even a perfect signal cannot reach \(\delta\)
(degenerate \(r\) or \(d\) near 0 on the content margin), the floor is
unreachable and the cell cannot support gating by construction; on the
squeeze plane (Figure 5) such cells are placed at signal surplus
\(\mathrm{AUROC}-1\).

\subsection{4.9 Extraction-completeness admission gate and the two
prompt
regimes}\label{extraction-completeness-admission-gate-and-the-two-prompt-regimes}

Auditing an early full-grid pass found that a substantial fraction of
samples in most cells had \texttt{a0\_answer}/\texttt{yf\_answer} equal
to \texttt{None} (generation truncated before an explicit answer marker)
or a literally echoed placeholder token, both silently scored as
incorrect. Root cause: the task/review prompts specified a bracketed
placeholder
(\texttt{\textquotesingle{}\#\#\#\#\ \textless{}answer\textgreater{}\textquotesingle{}})
that models sometimes echoed verbatim instead of substituting a value,
and the revision prompt invited a full re-derivation that regularly
exhausted the token budget before reaching the marker. Two responses:

\begin{enumerate}
\def\labelenumi{\arabic{enumi}.}
\tightlist
\item
  \textbf{An admission gate.} A cell passes iff both the \(a_0\)-bad and
  \(y_f\)-bad rate are \(<0.25\) (``bad'' = \texttt{None} or echoed
  placeholder; \texttt{cf\_core.extraction\_gate\_passes}, audited via
  \texttt{code/check\_extraction\_gate.py}). 29 of 31 generated cells
  pass; the two failures (both signal-D subsamples) are excluded
  throughout.
\item
  \textbf{A repaired prompt regime.} Concrete worked-example prompts in
  place of bracketed placeholders, an explicitly terse revision
  instruction, and larger token budgets. All 29 admitted cells were
  generated (or regenerated) under the repaired regime. The
  truncation-prone originals of eight cells are retained as a controlled
  before/after comparison --- the same (model, task) pairs under both
  regimes --- which provides C1's sign-reversal evidence (§9.1).
\end{enumerate}

A parallel bug in the MCQ extractor's permissive third fallback
(matching any standalone letter anywhere in truncated reasoning prose)
was fixed at the same time; truncation-regime MCQ cells' extracted
answers are unreliable even when not flagged \texttt{None}, which the
probe's fidelity control quantifies directly (§9.1).

\subsection{4.10 Forced-continuation
probe}\label{forced-continuation-probe}

For every admitted cell with \(\geq 10\) \(a_0\)-bad rows, we force a
same-turn greedy continuation of the original round-0 response ---
\texttt{{[}task\ prompt{]}\ +\ a0\_text\ +\ "Therefore,\ my\ final\ answer\ is"}
(32 tokens, no new user turn, no invitation to re-reason) --- and parse
the result with the task's extractor plus a fallback that also
recognizes the model's own
\texttt{\textquotesingle{}\#\#\#\#\ X\textquotesingle{}} convention when
it reproduces it inside the continuation. If the model already held the
answer and merely failed to emit it in the expected format, this cheap
probe should recover it at roughly the rate the full (expensive,
two-round) revision does. A fidelity control --- the same probe applied
to a random \(n=40\) subsample of \(a_0\)-OK rows --- must reproduce the
already-extracted \(a_0\) answer \(\geq 90\%\) of the time, or the
probe's own parsing (not the model) is doing the work. Decision rule per
cell: probe accuracy on \(a_0\)-bad rows \(\geq 0.8\times\) the
revision's own accuracy on those rows, \textbf{and} fidelity
\(\geq 0.90\). These probes were run against reconstructed pre-fix
prompts, and the
\texttt{\textquotesingle{}\#\#\#\#\ X\textquotesingle{}} fallback was
added during an offline reparse after inspecting parser failures. We
therefore use this arm as exploratory mechanistic evidence, not as an
independently confirmatory test.

\subsection{4.11 Paraphrase arm}\label{paraphrase-arm}

Two additional review-template wordings --- \texttt{v2} (explicitly
invites full re-derivation, evaluating each option) and \texttt{v3}
(terse, at-most-two-sentence justification) --- alongside the frozen
\texttt{v1}, run on 4B GSM8K and 4B MMLU (\(n=200\), generation only).
If C1 is right about mechanism, wording that increases truncation risk
(\texttt{v2}, longer) should move the format margin while leaving the
content margin roughly fixed; wording alone should not manufacture
genuine reconsideration.

\subsection{4.12 Constrained-decoding causal
control}\label{constrained-decoding-causal-control}

The recover/loss categories of §3.3 are observational: an initial answer
that fails to parse and a revision that does parse are consistent with a
purely cosmetic fix, but they do not rule out a genuine change of mind
that happens to arrive alongside a format fix. To move from
correlational to causal evidence, we hold the already-generated
\(a_0\_\mathrm{text}\) and \(a_f\_\mathrm{text}\) fixed for every row of
every admitted cell with a meaningful extraction channel and re-extract
the final answer via \textbf{grammar-constrained decoding} instead of
free text plus regex, guaranteeing parseability by construction.
\textbf{Precisely what this identifies:} the intervention causally tests
how much of the apparent total effect is removed by guaranteeing
structured, parseable output on top of reasoning the model already
produced --- it does not, and cannot, identify whether the model's
original free-form generation already carried the same latent answer,
because forcing a constrained continuation is itself a new (if minimal)
elicitation event, not a passive read-out of the frozen text. We use
``causal control'' in this narrower sense throughout:

\begin{Shaded}
\begin{Highlighting}[]
\NormalTok{MCQ (ARC, MMLU, CommonsenseQA, TruthfulQA{-}MC1):}
\NormalTok{  append "Final answer (a single letter):" to [prompt + a0\_text or yf\_text]}
\NormalTok{  force exactly 1 token from \{valid option letters\}, both bare (\textquotesingle{}A\textquotesingle{}) and}
\NormalTok{  space{-}prefixed (\textquotesingle{} A\textquotesingle{}) token forms allowed (both are single tokens in the}
\NormalTok{  Qwen tokenizer; verified empirically)}

\NormalTok{Open, numeric (GSM8K):}
\NormalTok{  append "Therefore, the final numerical answer (digits only) is"}
\NormalTok{  force up to 8 tokens from \{0{-}9, \textquotesingle{}.\textquotesingle{}, \textquotesingle{},\textquotesingle{}, \textquotesingle{}{-}\textquotesingle{}\} plus a leading{-}space token}
\NormalTok{  and EOS/newline as explicit stop options (Qwen tokenizes multi{-}digit}
\NormalTok{  numbers one digit per token, e.g. \textquotesingle{}123\textquotesingle{} {-}\textgreater{} [\textquotesingle{}1\textquotesingle{},\textquotesingle{}2\textquotesingle{},\textquotesingle{}3\textquotesingle{}], so this small}
\NormalTok{  alphabet is exhaustive for numeric answers)}
\end{Highlighting}
\end{Shaded}

Both grammars are implemented as a \texttt{logits\_processors} callback
(\texttt{mlx\_lm.generate.generate\_step}) that adds \(-\infty\) to
every disallowed vocabulary entry at the forced positions; the preceding
free-form CoT is generated exactly as in the main study and is never
constrained. Not run on MATH (answer space includes fractions, radicals,
and algebraic expressions too rich for a small character-class grammar)
or TriviaQA (free-text entity answers are not enumerable); scoped to the
14 admitted cells built on GSM8K or an MCQ task, which between them
include the two largest apparent gains in the grid (4B/9B MMLU). If C1
is right, forcing \textasciitilde100\% parseability should collapse
\(\Delta_{\mathrm{format}}\) toward zero and pull
\(\Delta_{\mathrm{total}}\) toward the already-reported
\(\Delta_{\mathrm{content}}\); because the forced continuation is itself
a fresh (if minimal) elicitation, a residual gap does not by itself
falsify C1, but a residual that fails to shrink at all, or that moves in
the wrong direction, would.

\subsection{4.13 Grader sensitivity
(MATH)}\label{grader-sensitivity-math}

The primary MATH grader is a string normalizer
(\texttt{run\_stage2.normalize\_math\_answer}), not a
symbolic-equivalence checker, and is a stated limitation. We re-grade
every \texttt{bothok} row of the three admitted MATH cells (4B, 9B,
Gemma-4-12B) with a computer-algebra grader
(\texttt{sympy.parsing.latex.parse\_latex} + \texttt{simplify}, falling
back to the string verdict when either side fails to parse as LaTeX) and
compare \(\Delta_{\mathrm{content}}\) under both graders. This tests
whether the string grader's known conservatism could be inflating the
format margins we attribute to C1, or is instead a symmetric,
content-margin-neutral measurement error.

\subsection{4.14 Literature protocol replication
(IoE)}\label{literature-protocol-replication-ioe}

We replicate Li et al.'s IoE prompting protocol \citep{li2024confidence}
verbatim from its released implementation
(\texttt{github.com/MBZUAI-CLeaR/IoE-Prompting},
\texttt{run\_math\_IoE.py}, fetched 2026-07-20), including its exact
prompt wording, its exact extraction convention and regex, and its
conditional third round:

\begin{Shaded}
\begin{Highlighting}[]
\NormalTok{Q1: question + " Explain your reasoning step{-}by{-}step." + EXTRACTOR}
\NormalTok{Q2 (IoE): "Review your previous answer. If you are confident about your}
\NormalTok{    answer, maintain your answer. Otherwise, update your answer." + EXTRACTOR}
\NormalTok{Q3 (Decision, only if P1\_ans != P2\_ans): "You give two different answers in}
\NormalTok{    previous responses. Check the problem and your answers again, and give}
\NormalTok{    the best answer." + EXTRACTOR}
\NormalTok{EXTRACTOR: " Your final answer should be put between two \#\#, like \#\# 1 \#\#}
\NormalTok{    (if your final answer is 1), at the end of your response."}
\end{Highlighting}
\end{Shaded}

on GSM8K (\(n=400\)) with Qwen3.5-4B/9B, greedy decoding, no thinking
mode. Their reported headline result is for \texttt{gpt-3.5-turbo-0613};
that exact snapshot was permanently retired by OpenAI on September 13,
2024 and cannot be obtained by any account today, so a literal
reproduction of their reported numbers is not possible for anyone. We
therefore run their protocol, unmodified, on a model family they did not
test, and ask whether the same margin-decomposition story holds --- a
protocol-fidelity replication, not a literal reproduction.

\subsection{4.15 Frontier family check via
API}\label{frontier-family-check-via-api}

The largest model locally hostable on this study's hardware (Apple
Silicon, 24GB unified memory) is Qwen3.5-9B; a 70B-class model at 4-bit
already exceeds available memory. To test whether the squeeze (C3)
persists, tightens, or opens at a scale genuinely beyond the primary
grid, we ran the identical always-revise protocol against two much
larger models via OpenRouter's free API tier: Tencent Hy3
(\texttt{tencent/hy3:free}) and Nvidia Nemotron-3-Ultra-550B
(\texttt{nvidia/nemotron-3-ultra-550b-a55b:free}, a
\textasciitilde55B-active-parameter MoE), on GSM8K/MMLU/MATH, \(n=200\)
target per task. This arm has two data-completeness caveats disclosed up
front: (i) free-tier API access is rate-limited, and OpenRouter retired
the Hy3 slug entirely partway through data collection (confirmed via
HTTP 404 ``unavailable for free'' on every subsequent request), so its
MATH file could not be collected at all and its MMLU file stopped at
153/200; Nemotron hit a hard daily quota mid-collection but was not
deprecated, and all three of its files (GSM8K/MMLU/MATH) were completed
to the full \(n=200\) target after the quota reset, across two
collection sessions roughly a day apart. (ii) Neither model returns
response logprobs on the free tier, so signals A/A\('\)/B/C are
unavailable for this arm; only the margin decomposition (which needs
only extracted answers and correctness) is reported. Four of the five
resulting cells fall outside the 25\% admission gate (§4.9) that governs
the primary 29-cell grid (Nemotron \(\times\) GSM8K narrowly clears it
at 22\% \(a_0\)-bad); this arm as a whole is reported as an exploratory
robustness check rather than folded into the admitted-cell pool, even
for the one cell that technically qualifies.

\section{5. Metrics}\label{metrics}

\subsection{5.1 Identity and
decomposition}\label{identity-and-decomposition}

\begin{Shaded}
\begin{Highlighting}[]
\NormalTok{helpful(tau) = P(trigger, y\_0=0, y\_f=1)}
\NormalTok{harmful(tau) = P(trigger, y\_0=1, y\_f=0)}
\NormalTok{Delta(tau)   = helpful {-} harmful   \# algebraically equal to SS3.2\textquotesingle{}s identity;}
\NormalTok{                                   \# restricted to bothok rows for all content{-}margin use}
\end{Highlighting}
\end{Shaded}

\subsection{5.2 Gating outcomes}\label{gating-outcomes}

\begin{Shaded}
\begin{Highlighting}[]
\NormalTok{gain   = acc\_select(tau*) {-} max(acc\_never, acc\_always)}
\NormalTok{regret = acc\_oracle {-} acc\_select(tau*)}
\NormalTok{regret rate = regret / (acc\_oracle {-} max(acc\_never, acc\_always))   \# target \textless{} 50\%}
\end{Highlighting}
\end{Shaded}

\subsection{5.3 Leave-one-out floor
prediction}\label{leave-one-out-floor-prediction}

Across sealed cells, the floor criterion's point prediction of holdout
\(\max_\tau \Delta\) (binormal \(\max\Delta\) at the cell's dev AUROC
and content-margin \((\mathrm{acc}_0, r, d)\)) is compared under
leave-one-out against two trivial baselines --- the constant median
predictor and a linear \(\mathrm{acc}_0\)-only fit --- on MAE, plus a
binary floor-pass confusion table (predicted
\(\mathrm{AUROC} > \mathrm{AUROC}^*\) versus actual holdout
\(\max\Delta \geq \delta\)). Implementation:
\texttt{cf\_core.loo\_floor\_predict}, driven by
\texttt{code/loo\_analysis.py}. Results in §9.4.

\subsection{5.4 Validation of the instrument on a synthetic
sandbox}\label{validation-of-the-instrument-on-a-synthetic-sandbox}

Before any of §4 is pointed at a real model, the instrument must do what
it claims on data with a known generating process. Six checks; all
reproducible from \texttt{code/cf\_core.py},
\texttt{code/sim\_trajectory.py}, \texttt{code/sim\_validate.py}, and
\texttt{code/make\_validation\_figures.py} (NumPy/SciPy only; no GPU, no
LLM). These checks validate the identity and floor apparatus of §3.2 in
isolation; the margin decomposition is a bookkeeping layer applied
\emph{before} the identity, not a change to the identity itself.

\textbf{Check 1: per-\(\tau\) identity to machine precision.} On a
synthetic always-revise population (\(n = 5000\)), we compute
\(\Delta(\tau)\) two ways at 81 thresholds: (a) helpful \(-\) harmful
from per-sample labels, and (b)
\((1-\mathrm{acc}_0)\cdot\mathrm{TPR}\cdot r - \mathrm{acc}_0\cdot\mathrm{FPR}\cdot d\).
The maximum absolute discrepancy is \textbf{\(< 10^{-12}\)} (Figure 1,
left).

\textbf{Check 2: ALWAYS endpoint and Huang localization.}
\(\Delta_{\mathrm{ALWAYS}} = (1-\mathrm{acc}_0)\cdot r - \mathrm{acc}_0\cdot d\)
is positive iff
\(\lambda = \mathrm{acc}_0\cdot d / ((1-\mathrm{acc}_0)\cdot r) < 1\).
Three analytic \((\mathrm{acc}_0, r, d)\) cases match the sign
prediction in all cases --- including a Huang-like open-reasoning point
(\(\mathrm{acc}_0 = 0.70\), \(r = 0.40\), \(d = 0.90\),
\(\lambda = 2.10\), \(\Delta_{\mathrm{ALWAYS}} < 0\)) where ALWAYS must
fail regardless of AUROC (Figure 1, right).

\begin{figure}
\centering
\pandocbounded{\includegraphics[keepaspectratio,alt={Figure 1: Left --- \textbackslash Delta(\textbackslash tau) from helpful-harmful vs.~the identity formula; points on the diagonal. Right --- \textbackslash Delta\_\{\textbackslash mathrm\{ALWAYS\}\} vs.~\textbackslash lambda; profit only for \textbackslash lambda \textless{} 1.}]{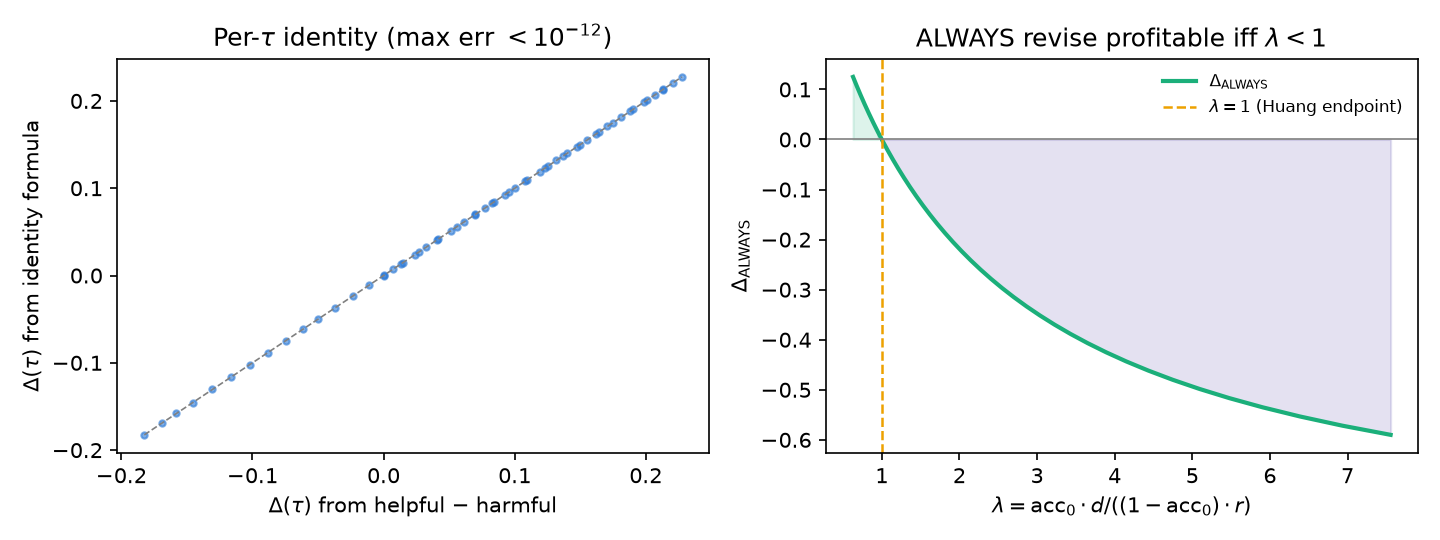}}
\caption{Figure 1: Left --- \(\Delta(\tau)\) from helpful\(-\)harmful
vs.~the identity formula; points on the diagonal. Right ---
\(\Delta_{\mathrm{ALWAYS}}\) vs.~\(\lambda\); profit only for
\(\lambda < 1\).}
\end{figure}

\textbf{Check 3: \(\delta\)-floor separates pass/fail cells.} On a
five-cell synthetic grid with known
\((\mathrm{acc}_0, r, d, \mathrm{AUROC})\), the floor
(\(\delta = 0.01\)) predicts whether
\(\max_\tau \Delta(\tau) \geq \delta\) with \textbf{100\%} agreement
(5/5). Figure 2 (right) shows measured AUROC vs.~solved AUROC*; green
points are floor-passing cells.

\textbf{Check 4: offline gating equals brute force.} For a held-out
\(\tau^*\), accuracy from the closed-form gate mask matches per-sample
\(\mathbb{1}[\mathrm{trigger}] \cdot y_f + \mathbb{1}[\neg\mathrm{trigger}] \cdot y_0\)
to \textbf{\(< 10^{-15}\)}.

\textbf{Check 5: explicit dominates implicit.} In \textbf{100\%} of
synthetic cells, the explicit confidence ROC lies above the implicit
flip point; the dominance gap correlates with gating gain over
max(ALWAYS, NEVER) at \(\rho = 0.99\) (Figure 3).

\begin{figure}
\centering
\pandocbounded{\includegraphics[keepaspectratio,alt={Figure 2: Left --- binormal ROCs vs.~the \textbackslash lambda floor ray for a high-d cell. Right --- measured AUROC vs.~\textbackslash delta-floor AUROC* across synthetic cells.}]{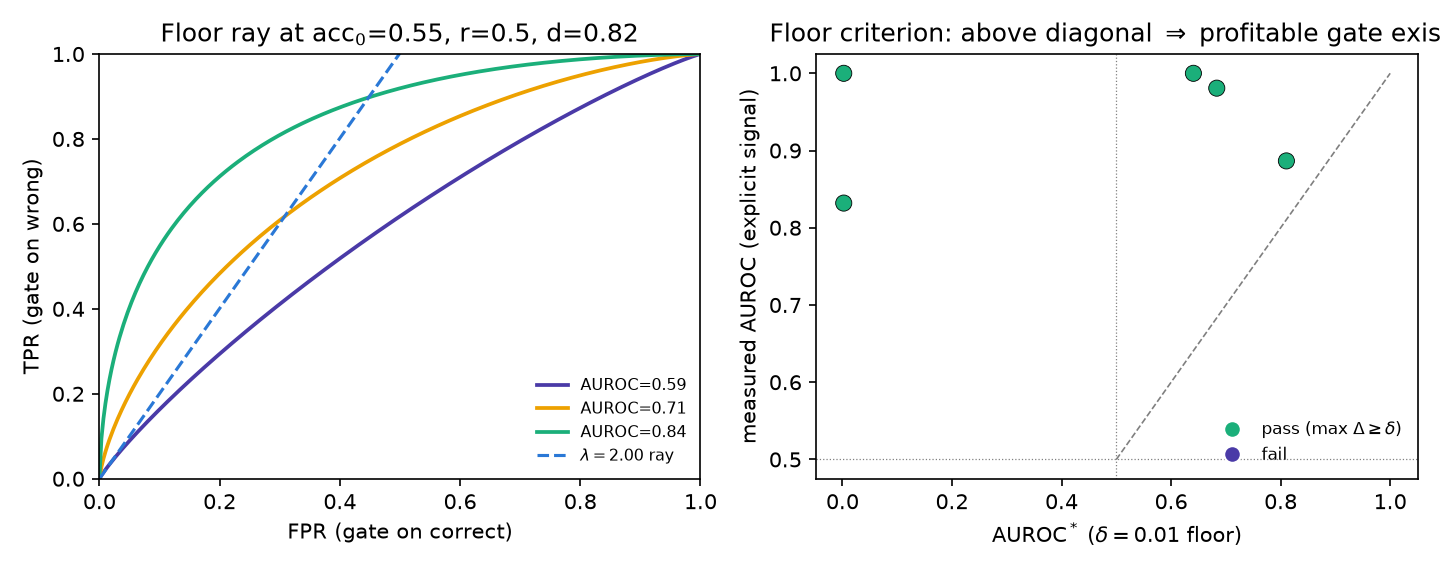}}
\caption{Figure 2: Left --- binormal ROCs vs.~the \(\lambda\) floor ray
for a high-\(d\) cell. Right --- measured AUROC vs.~\(\delta\)-floor
AUROC* across synthetic cells.}
\end{figure}

\begin{figure}
\centering
\pandocbounded{\includegraphics[keepaspectratio,alt={Figure 3: Left --- explicit ROC curves with implicit flip points (\textbackslash times) below. Right --- dominance gap vs.~gating gain across cells.}]{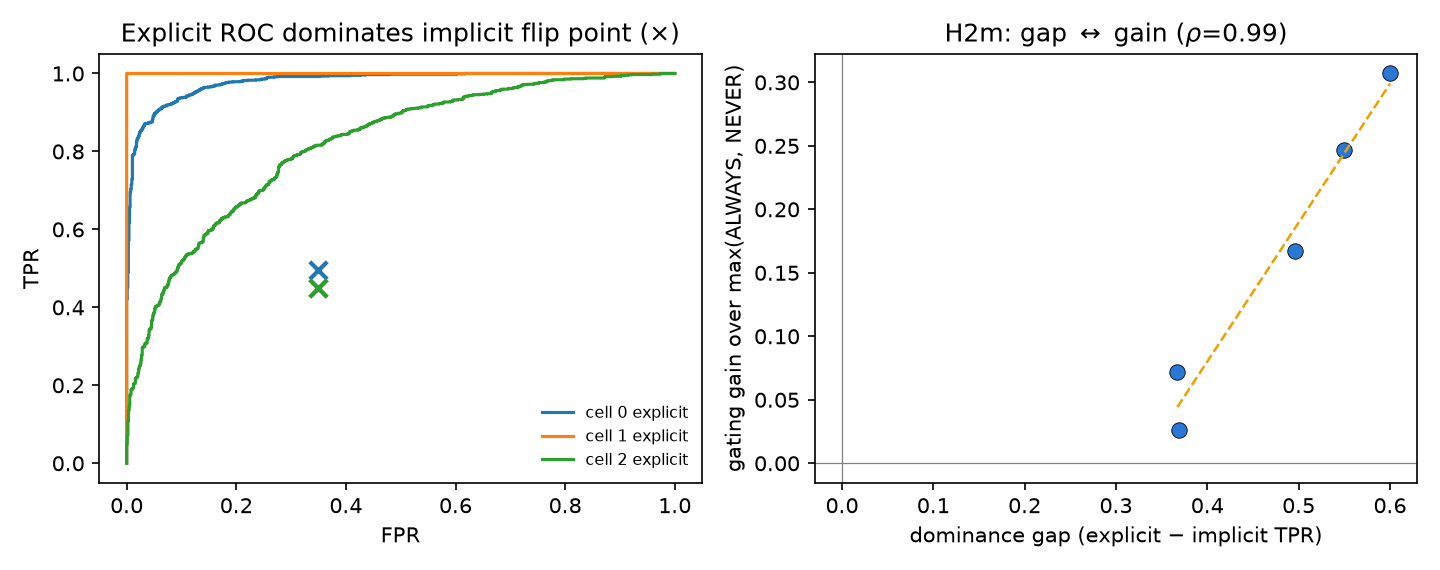}}
\caption{Figure 3: Left --- explicit ROC curves with implicit flip
points (\(\times\)) below. Right --- dominance gap vs.~gating gain
across cells.}
\end{figure}

\textbf{Check 6: \(\delta\) correction blocks the degenerate corner.} At
\(\mathrm{acc}_0 = 0.74\), \(r = 0.28\), \(d = 0.96\)
(\(\lambda = 9.76\)), a binormal signal with AUROC \(= 0.80\) still has
\(\max_\tau \Delta = 0.0026 < \delta\). Nominal ROC--\(\lambda\)
crossing is insufficient; the \(\delta\)-floor is necessary.

None of these six checks claim anything about real language models. They
establish that the identity, floor criterion, and offline evaluator
measure what §3.2/§5 specify. §9 is where C1--C3 are tested on real
trajectories.

\section{6. Experimental Design}\label{experimental-design}

\subsection{6.1 The grid}\label{the-grid}

{\def\LTcaptype{none} % do not increment counter
\begin{longtable}[]{@{}lll@{}}
\toprule\noalign{}
Component & Cells & n / cell \\
\midrule\noalign{}
\endhead
\bottomrule\noalign{}
\endlastfoot
Primary 4B (incl.~paraphrase v2/v3) & 12 tasks/variants & 200--400 \\
Scale 9B & 8 tasks & 400 (164 for code) \\
Floor 0.8B/2B & 6 (3 tasks \(\times\) 2 scales) & 300 \\
Gemma-4-12B family check & 3 tasks & 200 \\
\textbf{Total, admitted} & \textbf{29 cells} & \\
Excluded (fail extraction gate) & 2 (signal-D subsamples) & --- \\
\end{longtable}
}

Of the 29 admitted cells, 25 run the full sealed dev/holdout protocol;
the 4 paraphrase-arm cells are generation-only (no gating claim is made
on them). Eight truncation-regime originals are retained separately as
the §9.1 before/after comparison and are excluded from all primary
pools.

Post-review response arms, added after external review of this design
and analyzed against the already-admitted cells above rather than as new
admitted cells in their own right: the constrained-decoding causal
control (§4.12) covers 14 of the 29 admitted cells (re-extraction only,
no new admission decision); the IoE replication (§4.14) is 2 new
GSM8K-only trajectories (Qwen3.5-4B/9B, \(n=400\) each); the frontier
check (§4.15) is 5 trajectories on two additional model families
(Tencent Hy3, Nvidia Nemotron-3-Ultra-550B, GSM8K/MMLU/MATH, target
\(n=200\) each) that fail the 25\% admission gate and are reported as an
exploratory arm outside the 29-cell pool, not folded into it.

\subsection{6.2 Holdout isolation}\label{holdout-isolation}

Holdout labels are encrypted at generation time
(\texttt{code/evaluate\_holdout.py}: \texttt{seal} / \texttt{select-tau}
/ \texttt{decrypt-eval}); decryption occurs only after dev \(\tau^*\)
selection is frozen. This is a procedural safeguard against the
analyst's own temptation to peek: dev-phase \(\tau^*\) selection cannot
see holdout plaintext because the two halves live in separate files
until the final phase.

\subsection{6.3 Transparency and
provenance}\label{transparency-and-provenance}

This project was \textbf{not formally preregistered}: the Stage-1 design
document (identity, floor criterion, synthetic validation, task/model
grid; git tag \texttt{prereg-opt3-v2}) was frozen before real-model
generation but never publicly posted. The margin decomposition was
discovered on, and motivated by, an initial 16-cell grid; those cells
and the before/after prompt comparison are exploratory evidence. A dated
internal amendment (\texttt{ANALYSIS\_PLAN\_OPT3.md}, git tag
\texttt{prereg-opt3-v3-pivot}) specified the rank tests, regenerated/new
cells, paraphrase arm, probe rule, and Gemma family check before those
designated arms were completed. This is weaker than public
preregistration, several retained cells predate the amendment, and the
probe parser was subsequently repaired. We therefore describe the C1/C2
rank tests and post-pivot arms as specification-before-data evidence,
not as formally preregistered confirmation; C3, LOO, and single-cell
niche claims remain exploratory.

\section{7. Statistical Analysis}\label{statistical-analysis}

\begin{enumerate}
\def\labelenumi{\arabic{enumi}.}
\tightlist
\item
  \textbf{C1.} Row-level percentile bootstrap (\(B = 10^4\), seed 2026)
  of \(\Delta_{\mathrm{content}}\), \(\Delta_{\mathrm{format}}\),
  content-change-rate, and content-headroom per cell. Across cells with
  an active extraction channel (\(a_0\)-bad \(\geq 5\%\)), one-sided
  Wilcoxon signed-rank on
  \(|\Delta_{\mathrm{format}}| - |\Delta_{\mathrm{content}}|\). Probe
  decision rule per §4.10 with Wilson CIs. \textbf{Causal control
  (§4.12):} per cell, closure fraction
  \(= 1 - |\Delta_{\mathrm{total}}^{\mathrm{constrained}} - \Delta_{\mathrm{content}}| / |\Delta_{\mathrm{total}}^{\mathrm{original}} - \Delta_{\mathrm{content}}|\),
  computed only for cells with \(a_0\)-bad \(\geq 8\%\) under the
  original (unconstrained) extraction, to avoid dividing by a near-zero
  denominator on cells where format and content estimates were already
  close by construction.
\item
  \textbf{C2.} Per-cell bootstrap CI on \(\Delta_{\mathrm{content}}\)
  and content-change-rate against the inertia bounds (\(\pm0.03\),
  \(\leq 0.05\)). Floor-vs-capable contrast restricted to task families
  present at both scales (GSM8K/MMLU/ARC, to avoid confounding scale
  with task-geometry mix); one-sided Mann--Whitney U on change-rate and
  content-harm-rate at the \textbf{cell} level (6 vs.~6), supplemented
  by a \textbf{row-level clustered model}: logistic GEE (exchangeable
  working correlation, clustered by cell, controlling for task) on
  content-change and content-harm indicators across all \texttt{bothok}
  rows of the 12 matched-task cells, addressing the cell-level test's
  small effective \(n\) and the fact that cells sharing a checkpoint or
  task family are not independent replications.
\item
  \textbf{C3 (exploratory).} Content-margin AUROC vs.~content-margin
  AUROC* per sealed cell (dev split only), using the dev-best of five
  signals; bootstrap CI on content headroom; the squeeze plane of Figure
  5. Separately, sealed-holdout gating outcomes (gain, regret,
  random-gate comparison) use the prespecified \(p_{\mathrm{norm}}\)
  policy and the ordinary benchmark margin. \textbf{Floor redo:} for the
  six floor cells, in addition to the binormal AUROC* criterion, we (i)
  fit the binormal model's \(\mu\) to the empirical AUROC and check
  goodness-of-fit against the empirical ROC (max absolute TPR deviation,
  \(R^2\)) and (ii) compute a fully nonparametric floor-pass call ---
  percentile bootstrap (\(B=10^4\)) CI on the empirical
  \(\max_\tau \Delta(\tau)\) (direct threshold sweep, no distributional
  assumption), passing only if the CI lower bound clears \(\delta\).
\item
  \textbf{Paraphrase.} Paired by \texttt{task\_id} across the three
  review-template variants; format-margin ratio (max/min
  \(|\Delta_{\mathrm{format}}|\) across variants) and content-margin
  stability (\(\max\) deviation from \texttt{v1} \(\leq 0.02\)).
\item
  \textbf{LOO.} §5.3's leave-one-out comparison over the 25 sealed
  cells.
\item
  \textbf{Multiplicity.} Benjamini--Hochberg FDR \(0.05\) within the
  C1/C2 rank-test family (the C1 Wilcoxon test and the two C2
  Mann--Whitney tests); C3, LOO, the GEE model, and the floor redo are
  reported as CI/threshold/model-based comparisons, not part of the
  FDR-controlled family.
\end{enumerate}

All code: \texttt{code/stats\_tests.py},
\texttt{code/decompose\_margins.py}, \texttt{code/probe\_extraction.py},
\texttt{code/loo\_analysis.py}, \texttt{code/constrained\_probe.py},
\texttt{code/analyze\_constrained.py},
\texttt{code/cas\_grader\_sensitivity.py}, \texttt{code/floor\_redo.py},
\texttt{code/gee\_scale\_contrast.py},
\texttt{code/run\_ioe\_replication.py},
\texttt{code/run\_frontier\_api.py}. Outputs:
\texttt{data/stats\_tests.json},
\texttt{data/margin\_decomposition.json},
\texttt{data/loo\_floor\_prediction.json},
\texttt{data/floor\_redo.json}, \texttt{data/constrained/*.jsonl}.

\section{8. Compute Budget (Apple Silicon Mac
Mini)}\label{compute-budget-apple-silicon-mac-mini}

Realized throughput (4-bit MLX): 4B \textasciitilde17--32 s/sample
depending on task; 9B \textasciitilde1.9x the 4B rate, consistent across
matched tasks; floor-scale (0.8B/2B) \(3\)--\(14\) s/sample. Gemma-4-12B
via \texttt{ollama}: \textasciitilde41--61 s/sample. Total wall-clock
for probe + regeneration + paraphrase + family arms \textasciitilde40 h
on a single Mac Mini. Zero API cost through the primary 29-cell grid.

The post-review arms add: the constrained-decoding causal control
(§4.12), a forced short continuation per row reusing already-generated
text, \textasciitilde1--2 s/row, well under an hour in aggregate across
14 cells; the IoE replication (§4.14), full always-revise generation at
the same MLX throughput as above, \textasciitilde4 h/model; and the
frontier check (§4.15), OpenRouter's free tier, which introduces the
study's only non-zero external dependency --- not monetary cost, but
availability risk. That risk materialized: Hy3's free-tier listing was
retired by OpenRouter mid-collection, and both models were subject to
per-day request quotas that left several task files incomplete (§9.8
reports exact \(n\) per file). Wall-clock for the frontier arm was
dominated by queueing and retry backoff rather than generation,
\textasciitilde8--36 h/model/task depending on rate-limit pressure at
the time.

\section{9. Results}\label{results}

Figure 4 shows the complete margin decomposition for all 29 admitted
cells; the numeric table is §9.6. The visual pattern \emph{is} the
paper's first claim: the amber format-recovery component, not the blue
content component, carries nearly every large
\(\Delta_{\mathrm{total}}\).

\begin{figure}
\centering
\pandocbounded{\includegraphics[keepaspectratio,alt={Figure 4: Margin decomposition of the ALWAYS-revise effect for all 29 admitted cells. Signed stacked bars decompose \textbackslash Delta\_\{\textbackslash mathrm\{total\}\} (black dot) exactly into \textbackslash Delta\_\{\textbackslash mathrm\{content\}\} (blue), \textbackslash Delta\_\{\textbackslash mathrm\{format\textbackslash text\{-\}recover\}\} (amber), and \textbackslash Delta\_\{\textbackslash mathrm\{format\textbackslash text\{-\}loss\}\} (purple). Large apparent gains (9B MMLU +0.145, 4B MMLU +0.105, MATH at both scales) are almost purely amber; genuinely negative content effects (blue, leftward) appear only at floor scale.}]{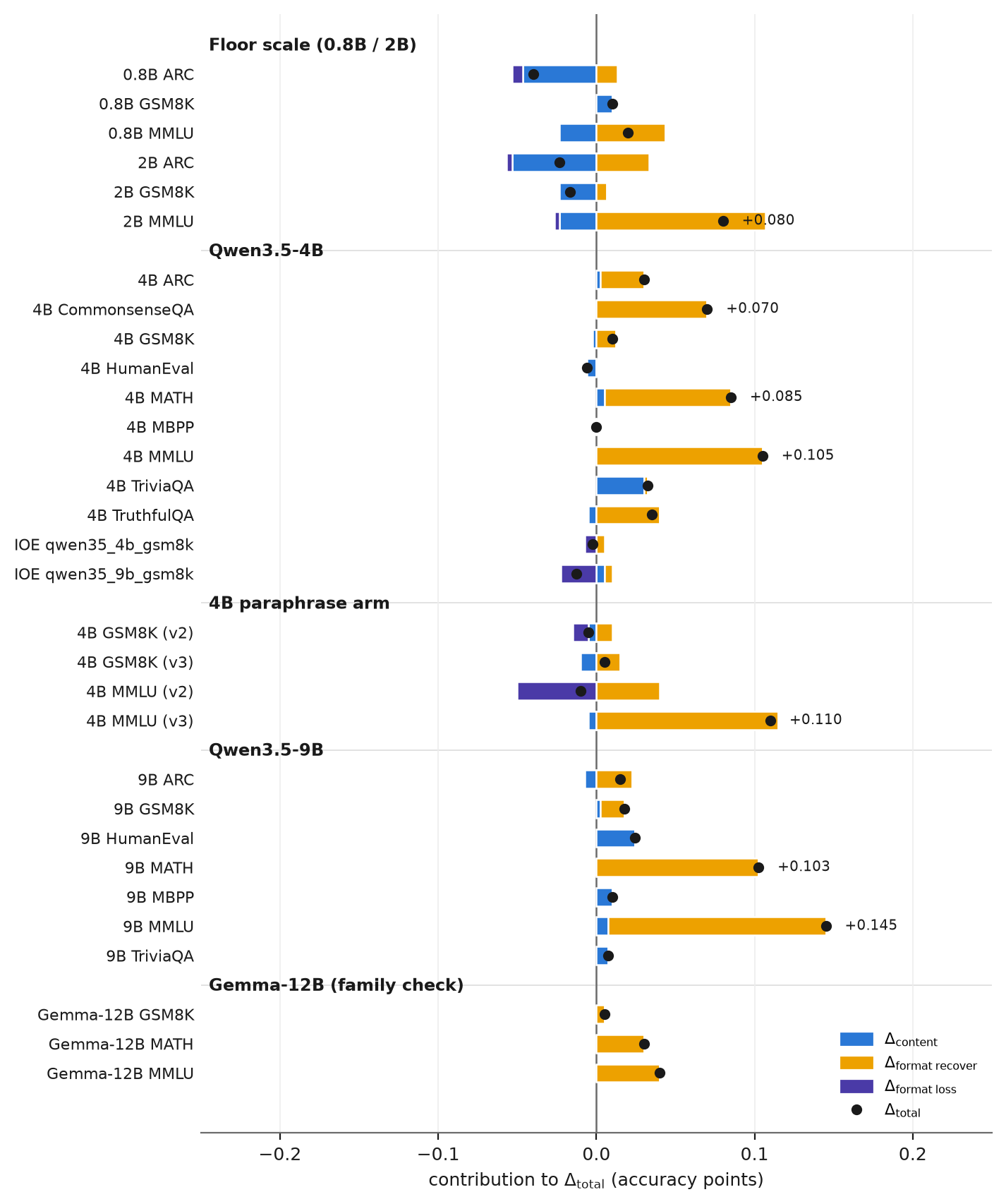}}
\caption{Figure 4: Margin decomposition of the ALWAYS-revise effect for
all 29 admitted cells. Signed stacked bars decompose
\(\Delta_{\mathrm{total}}\) (black dot) exactly into
\(\Delta_{\mathrm{content}}\) (blue),
\(\Delta_{\mathrm{format\text{-}recover}}\) (amber), and
\(\Delta_{\mathrm{format\text{-}loss}}\) (purple). Large apparent gains
(9B MMLU \(+0.145\), 4B MMLU \(+0.105\), MATH at both scales) are almost
purely amber; genuinely negative content effects (blue, leftward) appear
only at floor scale.}
\end{figure}

\subsection{9.1 C1: format artifacts dominate, and they dominate in
both
directions}\label{c1-format-artifacts-dominate-and-they-dominate-in-both-directions}

The sharpest exploratory evidence is a same-cell sign reversal between
the truncation-prone and repaired prompt regimes (§4.9):

{\def\LTcaptype{none} % do not increment counter
\begin{longtable}[]{@{}
  >{\raggedright\arraybackslash}p{(\linewidth - 8\tabcolsep) * \real{0.1579}}
  >{\raggedleft\arraybackslash}p{(\linewidth - 8\tabcolsep) * \real{0.2105}}
  >{\raggedleft\arraybackslash}p{(\linewidth - 8\tabcolsep) * \real{0.2105}}
  >{\raggedleft\arraybackslash}p{(\linewidth - 8\tabcolsep) * \real{0.2105}}
  >{\raggedleft\arraybackslash}p{(\linewidth - 8\tabcolsep) * \real{0.2105}}@{}}
\toprule\noalign{}
\begin{minipage}[b]{\linewidth}\raggedright
Cell
\end{minipage} & \begin{minipage}[b]{\linewidth}\raggedleft
truncation-regime \(\Delta_{\mathrm{total}}\)
\end{minipage} & \begin{minipage}[b]{\linewidth}\raggedleft
its \(\Delta_{\mathrm{format}}\)
\end{minipage} & \begin{minipage}[b]{\linewidth}\raggedleft
repaired \(\Delta_{\mathrm{total}}\)
\end{minipage} & \begin{minipage}[b]{\linewidth}\raggedleft
its \(\Delta_{\mathrm{format}}\)
\end{minipage} \\
\midrule\noalign{}
\endhead
\bottomrule\noalign{}
\endlastfoot
0.8B \(\times\) ARC & \(+0.007\) & \(-0.037\) & \(-0.040\) &
\(+0.007\) \\
0.8B \(\times\) GSM8K & \(-0.090\) & \(-0.103\) & \(+0.010\) &
\(+0.000\) \\
0.8B \(\times\) MMLU & \(+0.000\) & \(-0.007\) & \(+0.020\) &
\(+0.043\) \\
2B \(\times\) ARC & \(-0.137\) & \(-0.133\) & \(-0.023\) & \(+0.030\) \\
2B \(\times\) GSM8K & \(-0.213\) & \(-0.123\) & \(-0.017\) &
\(+0.007\) \\
2B \(\times\) MMLU & \(-0.053\) & \(-0.057\) & \(+0.080\) &
\(+0.103\) \\
4B \(\times\) ARC & \(+0.080\) & \(+0.058\) & \(+0.030\) & \(+0.028\) \\
4B \(\times\) GSM8K & \(-0.068\) & \(-0.065\) & \(+0.010\) &
\(+0.013\) \\
\end{longtable}
}

In every row, \(\Delta_{\mathrm{format}}\) tracks
\(\Delta_{\mathrm{total}}\) far more closely than
\(\Delta_{\mathrm{content}}\), and the between-regime swing is
concentrated in the format margin. Prompt changes can in principle alter
reasoning as well as formatting, so this comparison alone is not causal
identification. Its diagnostic value is that the measured content margin
remains small while extraction failures move from the revision side
(apparent harm) to the initial-answer side (apparent gain).

Across the 12 admitted cells with an active extraction channel
(\(a_0\)-bad \(\geq 5\%\)), \(|\Delta_{\mathrm{format}}|\) exceeds
\(|\Delta_{\mathrm{content}}|\) with median difference \(+0.055\);
one-sided Wilcoxon signed-rank \(p = 1.7\times10^{-3}\) (BH-adjusted
within the C1/C2 family, \(2.6\times10^{-3}\)).

\textbf{Forced-continuation probe (§4.10).} Ten cells had \(\geq 10\)
\(a_0\)-bad rows to probe. Two (both truncation-regime ARC) fail the
fidelity control outright (0.28--0.43 vs.~the 0.90 threshold) --- itself
evidence: it quantifies how often the pre-fix MCQ extractor's
``successful'' letter reads were wrong even when not flagged
\texttt{None} (§4.9). Of the eight fidelity-clean cells, six satisfy the
decision rule:

{\def\LTcaptype{none} % do not increment counter
\begin{longtable}[]{@{}
  >{\raggedright\arraybackslash}p{(\linewidth - 8\tabcolsep) * \real{0.1667}}
  >{\raggedleft\arraybackslash}p{(\linewidth - 8\tabcolsep) * \real{0.2222}}
  >{\raggedleft\arraybackslash}p{(\linewidth - 8\tabcolsep) * \real{0.2222}}
  >{\raggedleft\arraybackslash}p{(\linewidth - 8\tabcolsep) * \real{0.2222}}
  >{\raggedright\arraybackslash}p{(\linewidth - 8\tabcolsep) * \real{0.1667}}@{}}
\toprule\noalign{}
\begin{minipage}[b]{\linewidth}\raggedright
Cell
\end{minipage} & \begin{minipage}[b]{\linewidth}\raggedleft
probe acc (95\% CI)
\end{minipage} & \begin{minipage}[b]{\linewidth}\raggedleft
\(0.8\times\) revision acc
\end{minipage} & \begin{minipage}[b]{\linewidth}\raggedleft
fidelity
\end{minipage} & \begin{minipage}[b]{\linewidth}\raggedright
rule
\end{minipage} \\
\midrule\noalign{}
\endhead
\bottomrule\noalign{}
\endlastfoot
4B GSM8K & 0.635 {[}0.50, 0.75{]} & 0.154 & 1.00 & \textbf{pass} \\
4B MATH & 0.378 {[}0.28, 0.49{]} & 0.346 & 1.00 & \textbf{pass} \\
4B TriviaQA & 0.083 {[}0.01, 0.35{]} & 0.067 & 0.98 & pass (n=12, wide
CI) \\
4B TruthfulQA-MC1 & 0.571 {[}0.37, 0.76{]} & 0.305 & 1.00 &
\textbf{pass} \\
9B MATH & 0.391 {[}0.29, 0.50{]} & 0.377 & 1.00 & pass (narrow) \\
9B MMLU & 0.658 {[}0.55, 0.75{]} & 0.557 & 1.00 & pass (narrow) \\
4B MMLU & 0.413 {[}0.31, 0.53{]} & 0.448 & 1.00 & fail \\
4B CommonsenseQA & 0.267 {[}0.14, 0.44{]} & 0.373 & 1.00 & fail \\
\end{longtable}
}

4B GSM8K is the sharpest case: a zero-reasoning forced continuation
recovers correct answers on 63.5\% of the rows that the full two-round
revision protocol only recovers 19.2\% of the time --- the expensive
revision does \emph{worse} than simply asking again for the answer,
which is only consistent with a format story. The two exceptions (4B
MMLU, 4B CommonsenseQA) are honest counter-examples: on these cells the
full revision recovers more than the probe, suggesting the revision turn
does more than pure format repair there even though the content margin
elsewhere is near zero (§9.2).

\textbf{Paraphrase arm (§4.11).} Holding the model and task fixed and
varying only the review template's wording:

{\def\LTcaptype{none} % do not increment counter
\begin{longtable}[]{@{}
  >{\raggedright\arraybackslash}p{(\linewidth - 6\tabcolsep) * \real{0.2308}}
  >{\raggedright\arraybackslash}p{(\linewidth - 6\tabcolsep) * \real{0.2308}}
  >{\raggedleft\arraybackslash}p{(\linewidth - 6\tabcolsep) * \real{0.3077}}
  >{\raggedright\arraybackslash}p{(\linewidth - 6\tabcolsep) * \real{0.2308}}@{}}
\toprule\noalign{}
\begin{minipage}[b]{\linewidth}\raggedright
Task
\end{minipage} & \begin{minipage}[b]{\linewidth}\raggedright
\(|\Delta_{\mathrm{format}}|\): v1 / v2 / v3
\end{minipage} & \begin{minipage}[b]{\linewidth}\raggedleft
ratio
\end{minipage} & \begin{minipage}[b]{\linewidth}\raggedright
\(\Delta_{\mathrm{content}}\): v1 / v2 / v3
\end{minipage} \\
\midrule\noalign{}
\endhead
\bottomrule\noalign{}
\endlastfoot
GSM8K & 0.013 / 0.000 / 0.015 & \(\geq 15\times\) & \(-0.003\) /
\(-0.005\) / \(-0.010\) \\
MMLU & 0.105 / 0.010 / 0.115 & \(12\times\) & \(+0.000\) / \(+0.000\) /
\(-0.005\) \\
\end{longtable}
}

\texttt{v2} (explicitly invites re-deriving each option) drives MMLU's
format margin from \(+0.105\) to \(-0.010\) --- a full sign flip ---
mechanistically because its longer instruction reintroduces the
truncation-before-marker failure mode. The content margin stays within
\(0.02\) of \texttt{v1} in all four variant/task combinations. Format is
sensitive to how the question is asked; content is not.

\textbf{Constrained-decoding causal control (§4.12).} All 14 cells reach
\(0\%\) \(a_0\)/\(a_f\)-bad under grammar-constrained re-extraction,
confirming the mechanism works as intended. The central comparison is
whether \(\Delta_{\mathrm{total}}\) under this forced-parseable
re-extraction moves toward the already-reported
\(\Delta_{\mathrm{content}}\):

{\def\LTcaptype{none} % do not increment counter
\begin{longtable}[]{@{}
  >{\raggedright\arraybackslash}p{(\linewidth - 10\tabcolsep) * \real{0.1304}}
  >{\raggedleft\arraybackslash}p{(\linewidth - 10\tabcolsep) * \real{0.1739}}
  >{\raggedleft\arraybackslash}p{(\linewidth - 10\tabcolsep) * \real{0.1739}}
  >{\raggedleft\arraybackslash}p{(\linewidth - 10\tabcolsep) * \real{0.1739}}
  >{\raggedleft\arraybackslash}p{(\linewidth - 10\tabcolsep) * \real{0.1739}}
  >{\raggedleft\arraybackslash}p{(\linewidth - 10\tabcolsep) * \real{0.1739}}@{}}
\toprule\noalign{}
\begin{minipage}[b]{\linewidth}\raggedright
Cell
\end{minipage} & \begin{minipage}[b]{\linewidth}\raggedleft
\(a_0\)-bad (original)
\end{minipage} & \begin{minipage}[b]{\linewidth}\raggedleft
\(\Delta_{\mathrm{total}}\) (original)
\end{minipage} & \begin{minipage}[b]{\linewidth}\raggedleft
\(\Delta_{\mathrm{content}}\)
\end{minipage} & \begin{minipage}[b]{\linewidth}\raggedleft
\(\Delta_{\mathrm{total}}\) (constrained)
\end{minipage} & \begin{minipage}[b]{\linewidth}\raggedleft
closure
\end{minipage} \\
\midrule\noalign{}
\endhead
\bottomrule\noalign{}
\endlastfoot
2B ARC & 6.3\% & \(-0.023\) & \(-0.053\) & \(-0.053\) & 100\% \\
4B ARC & 4.8\% & \(+0.030\) & \(+0.003\) & \(+0.003\) & 100\% \\
0.8B MMLU & 14.3\% & \(+0.020\) & \(-0.023\) & \(-0.017\) & 85\% \\
2B MMLU & 22.7\% & \(+0.080\) & \(-0.023\) & \(-0.003\) & 81\% \\
4B TruthfulQA-MC1 & 10.5\% & \(+0.035\) & \(-0.005\) & \(+0.005\) &
75\% \\
9B ARC & 4.0\% & \(+0.015\) & \(-0.008\) & \(-0.003\) & 78\% \\
9B GSM8K & 2.0\% & \(+0.018\) & \(+0.003\) & \(+0.008\) & 67\% \\
9B MMLU & 19.8\% & \(+0.145\) & \(+0.008\) & \(+0.053\) & 67\% \\
4B CommonsenseQA & 15.0\% & \(+0.070\) & \(0.000\) & \(+0.025\) &
64\% \\
4B MMLU & 18.8\% & \(+0.105\) & \(0.000\) & \(+0.048\) & 55\% \\
4B GSM8K & 3.0\% & \(+0.010\) & \(-0.003\) & \(-0.005\) & (low bad
rate) \\
0.8B ARC & 4.3\% & \(-0.040\) & \(-0.047\) & \(-0.037\) & (low bad
rate) \\
0.8B GSM8K & 3.3\% & \(+0.010\) & \(+0.010\) & \(+0.013\) & (low bad
rate) \\
2B GSM8K & 5.7\% & \(-0.017\) & \(-0.023\) & \(-0.027\) & (low bad
rate) \\
\end{longtable}
}

Restricting to the eight cells with a meaningfully active extraction
channel (\(a_0\)-bad \(\geq 8\%\) under the original extractor), median
closure is \(71\%\) (mean \(71\%\), range \(55\)--\(100\%\)); two cells
(2B/4B ARC) converge exactly. The four low-bad-rate cells are shown for
completeness but are not part of this comparison --- original, content,
and constrained estimates were already close together, and small
movements among them are noise, not evidence either way. This is the
paper's clearest causal result in the narrower sense defined in §4.12
--- freezing the model's reasoning text and only guaranteeing extraction
reliability at re-extraction closes most of the gap between the naive
total effect and the content-margin estimate on the cells where that gap
was largest, including a full sign flip on 0.8B MMLU
(\(+0.020 \to -0.017\)) and a change from the grid's single largest
apparent gain to less than half of it on 9B MMLU
(\(+0.145 \to +0.053\)). It is not, however, complete: 4B/9B MMLU retain
a residual \(+0.048\)/\(+0.053\) against a content-margin estimate near
zero. We see two candidate explanations and do not adjudicate between
them: either the imputation-based \(\Delta_{\mathrm{content}}\) (§3.3)
under-estimates a genuine content effect concentrated on
originally-unparseable rows, or the forced continuation itself --- a
fresh, if minimal, elicitation --- recovers slightly more than a purely
mechanical re-extraction would. Either way, the residual is far smaller
than the naive \(\Delta_{\mathrm{total}}\) these two cells originally
reported, and C1's qualitative claim (format is the primary driver)
survives; its quantitative claim (a specific, universal percentage) does
not, and we report the range rather than collapse it to one number.

\textbf{Grader sensitivity (§4.13).} Re-grading MATH with a CAS checker
instead of the string normalizer flips \(92/326\) (4B), \(86/313\) (9B),
and \(54/185\) (Gemma-4-12B) \texttt{bothok} rows from wrong to right
--- the string grader under-counts raw accuracy by roughly \(28\%\) of
these rows, and no row moves the other way. Despite this large swing in
raw accuracy, \(\Delta_{\mathrm{content}}\) is unchanged to three
decimal places under both graders on all three cells
(\(4\mathrm{B}: +0.0061 \to +0.0061\);
\(9\mathrm{B}, \mathrm{Gemma}: 0.0000 \to 0.0000\)), because the string
grader's conservatism applies symmetrically to \(a_0\) and \(a_f\). The
extraction/grading confound this paper is about is therefore not itself
an artifact of grader choice on the content margin, though it does mean
raw MATH accuracy numbers reported anywhere in this paper should be read
as conservative.

\subsection{9.2 C2: content inertia at capable scale, real flips at
floor
scale}\label{c2-content-inertia-at-capable-scale-real-flips-at-floor-scale}

Restricted to the content margin (\texttt{bothok} rows only), 11 of 13
capable-scale (4B/9B/12B) answer-level, non-ARC cells satisfy the
inertia bounds (\(\Delta_{\mathrm{content}}\) 95\% CI within
\(\pm0.03\), change rate \(\leq 0.05\)):

{\def\LTcaptype{none} % do not increment counter
\begin{longtable}[]{@{}
  >{\raggedright\arraybackslash}p{(\linewidth - 6\tabcolsep) * \real{0.2308}}
  >{\raggedright\arraybackslash}p{(\linewidth - 6\tabcolsep) * \real{0.2308}}
  >{\raggedleft\arraybackslash}p{(\linewidth - 6\tabcolsep) * \real{0.3077}}
  >{\raggedright\arraybackslash}p{(\linewidth - 6\tabcolsep) * \real{0.2308}}@{}}
\toprule\noalign{}
\begin{minipage}[b]{\linewidth}\raggedright
tier
\end{minipage} & \begin{minipage}[b]{\linewidth}\raggedright
representative cells
\end{minipage} & \begin{minipage}[b]{\linewidth}\raggedleft
content change rate
\end{minipage} & \begin{minipage}[b]{\linewidth}\raggedright
\(\Delta_{\mathrm{content}}\) 95\% CI
\end{minipage} \\
\midrule\noalign{}
\endhead
\bottomrule\noalign{}
\endlastfoot
capable, inert & 4B/9B MMLU, MATH, CommonsenseQA, TruthfulQA, GSM8K;
Gemma GSM8K/MMLU/MATH & \(0.000\)--\(0.009\) & contains 0 \\
capable, honest exceptions & 4B/9B TriviaQA & \(0.11\)--\(0.19\) &
excludes 0 (\(+0.01\) to \(+0.05\)) \\
capable, code (change-rate n/a for text-level edits) & 4B/9B
HumanEval/MBPP & --- & \(-0.03\) to \(+0.05\) \\
\end{longtable}
}

TriviaQA is a genuine, small, positive exception: short free-text
answers apparently do get real (if modest) benefit from a second look,
at both 4B and 9B. It reappears in §9.3 as the one marginal occupant of
the viable-gating region --- the two observations are the same
phenomenon.

Floor-scale cells (0.8B/2B) tell a different story on the matched task
families:

{\def\LTcaptype{none} % do not increment counter
\begin{longtable}[]{@{}lrr@{}}
\toprule\noalign{}
Cell & content change rate & \(\Delta_{\mathrm{content}}\) \\
\midrule\noalign{}
\endhead
\bottomrule\noalign{}
\endlastfoot
0.8B ARC & 0.172 & \(-0.047\) \\
0.8B MMLU & 0.199 & \(-0.023\) \\
2B GSM8K & 0.155 & \(-0.023\) \\
2B ARC & 0.075 & \(-0.053\) \\
2B MMLU & 0.074 & \(-0.023\) \\
0.8B GSM8K (exception) & 0.045 & \(+0.010\) \\
\end{longtable}
}

Five of six floor cells show real content-level instability (7.4--19.9\%
change rate, versus 0.0--3.1\% on the matched capable cells) with a
net-\emph{harmful} content effect. Matched-task contrast: change-rate
medians \(0.115\) (floor) vs.~\(0.005\) (capable), one-sided
Mann--Whitney \(p = 1.1\times10^{-3}\) (BH-adjusted
\(2.6\times10^{-3}\)); content-harm-rate medians \(0.055\)
vs.~\(0.001\), \(p = 1.7\times10^{-2}\) (BH-adjusted). 0.8B GSM8K is the
one exception (low change rate, small positive effect); numeric-answer
format may simply be more stable than MCQ reasoning even at floor scale,
but we do not have a confirmed mechanism and flag it as unresolved.

Because this 6-vs-6 cell-level test treats each cell as an independent
unit despite many rows within a cell coming from the same model/task
pair, we re-test with a row-level logistic GEE (exchangeable
correlation, clustered by cell, controlling for task; \(n=3815\) rows
across the 12 matched-task cells): floor-tier rows have \(15.9\times\)
higher odds of a content-level change (\(95\%\) CI \([5.8, 43.6]\),
\(p=8.2\times10^{-8}\)) and \(21.0\times\) higher odds of a harmful
content-level flip (\(95\%\) CI \([8.4, 52.6]\),
\(p=7.9\times10^{-11}\)) than capable-tier rows.

\textbf{What the GEE does and does not fix.} Clustering by cell corrects
for the within-cell row correlation that a naive row-level model would
ignore. It does not, by itself, address a separate non-independence: the
matched-task grid's 6 floor-tier and 6 capable-tier cells come from only
\textbf{4 distinct checkpoints} (0.8B, 2B, 4B, 9B) \(\times\) 3 tasks,
so they are not 12 independent draws either, and asymptotic GEE
\(p\)-values this small should not be read as if they were
(\texttt{code/tier\_contrast\_robustness.py}):

\begin{itemize}
\tightlist
\item
  \textbf{Task-paired analysis.} For each of the 3 matched tasks, the
  floor-tier checkpoints' (0.8B+2B) mean content-change/harmful rate
  exceeds the capable-tier checkpoints' (4B+9B) mean: all 3 of 3 tasks
  agree in direction for both outcomes (change-rate diffs \(+0.095\),
  \(+0.121\), \(+0.118\); harmful-rate diffs \(+0.023\), \(+0.060\),
  \(+0.072\)). A sign test across only 3 tasks floors at \(p=0.25\)
  regardless of effect size --- we report the floor rather than a
  smaller number, since no test can honestly claim more from 3 paired
  observations.
\item
  \textbf{Exact checkpoint-level permutation.} Pooling rows by
  checkpoint (collapsing across tasks) gives one change-rate and one
  harmful-rate per checkpoint. There are only 3 distinct ways to split 4
  checkpoints into two groups of 2, so an exact permutation test over
  checkpoint-tier assignment has a hard floor of \(p=1/3\) --- the
  observed floor-vs-capable split is among the most extreme of the 3,
  giving exact \(p=0.333\) for both outcomes. This is not a failure of
  the method; it is the honest ceiling of significance obtainable from 4
  checkpoints, and it is the number we report rather than the GEE's
  asymptotic \(p<10^{-7}\).
\item
  \textbf{Checkpoint-clustered bootstrap.} Resampling checkpoints (not
  rows) with replacement within each tier and recomputing the odds ratio
  each time (median of 2 resampled checkpoints per tier, \(B=10^4\))
  gives OR \(16.1\times\) {[}\(95\%\) CI \(7.7\), \(76.5\){]} for
  content change and OR \(19.4\times\) {[}\(11.4\), \(52.7\){]} for harm
  --- medians close to the GEE point estimates (\(15.9\times\),
  \(21.0\times\)) but with substantially wider intervals that still
  exclude \(1\), i.e.~still directionally decisive but honestly wider
  once checkpoint-level (not just row-level) resampling uncertainty is
  acknowledged.
\end{itemize}

Taken together, these checks support a real, consistent floor-vs-capable
contrast in the specific checkpoints tested, but not a claim stronger
than that: read it as ``a clear difference across the checkpoints
evaluated here,'' not a general scaling law, since model scale and
checkpoint identity are confounded by design (one checkpoint per scale
point).

\subsection{9.3 C3: the squeeze}\label{c3-the-squeeze}

Figure 5 provides an exploratory placement of all 25 sealed cells on the
(content headroom, signal surplus) plane. Both squeeze jaws are visible
at once: the floor-scale cells (green) sit right of the \(\delta\) line
with real headroom but \emph{below} the zero-surplus line --- their
dev-selected best signal cannot clear the content-margin floor --- while
nearly all capable cells (blue/amber) hug the left edge with essentially
no headroom regardless of signal quality.

\begin{figure}
\centering
\pandocbounded{\includegraphics[keepaspectratio,alt={Figure 5: The squeeze plane. x: content headroom (oracle - max(NEVER, ALWAYS) on both-parseable rows); y: best-signal AUROC minus the content-margin \textbackslash delta-floor AUROC* (dev split; unreachable floors plotted at AUROC - 1). The shaded quadrant --- real headroom and floor-passing signal --- contains exactly one marginal member, 9B TriviaQA. Cells pinned to the top-left (e.g.~4B TriviaQA, 9B HumanEval) have trivially-low floors (\textbackslash lambda \textbackslash to 0: ALWAYS never breaks anything, so no selective gate is needed) but no headroom for selection to exploit.}]{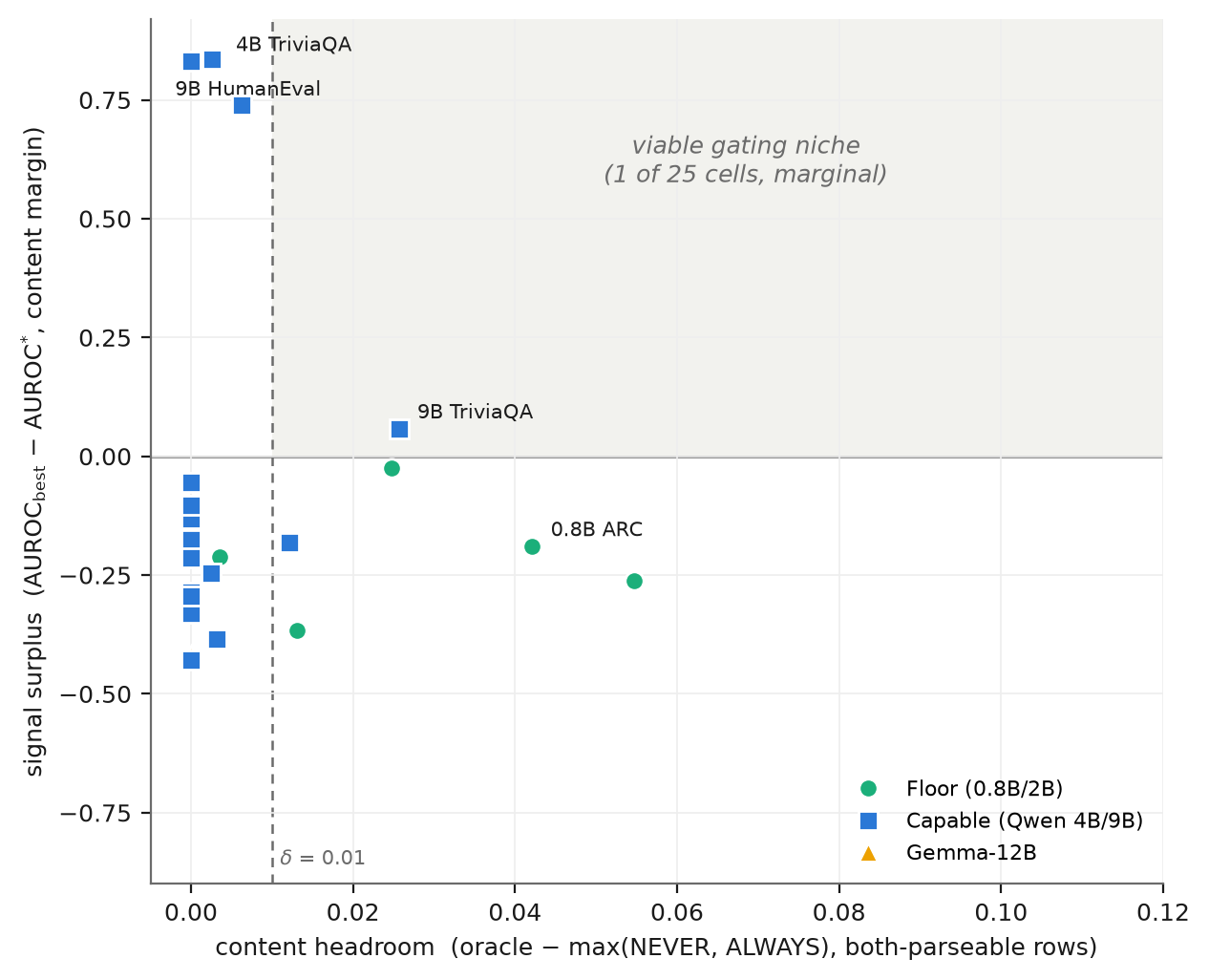}}
\caption{Figure 5: The squeeze plane. x: content headroom (oracle \(-\)
max(NEVER, ALWAYS) on both-parseable rows); y: best-signal AUROC minus
the content-margin \(\delta\)-floor AUROC* (dev split; unreachable
floors plotted at AUROC \(-\) 1). The shaded quadrant --- real headroom
\emph{and} floor-passing signal --- contains exactly one marginal
member, 9B TriviaQA. Cells pinned to the top-left (e.g.~4B TriviaQA, 9B
HumanEval) have trivially-low floors (\(\lambda \to 0\): ALWAYS never
breaks anything, so no \emph{selective} gate is needed) but no headroom
for selection to exploit.}
\end{figure}

\textbf{Floor side --- real headroom, insufficient signal.} All six
floor cells fail the content-margin floor under the binormal criterion:

{\def\LTcaptype{none} % do not increment counter
\begin{longtable}[]{@{}
  >{\raggedright\arraybackslash}p{(\linewidth - 10\tabcolsep) * \real{0.1364}}
  >{\raggedleft\arraybackslash}p{(\linewidth - 10\tabcolsep) * \real{0.1818}}
  >{\raggedright\arraybackslash}p{(\linewidth - 10\tabcolsep) * \real{0.1364}}
  >{\raggedleft\arraybackslash}p{(\linewidth - 10\tabcolsep) * \real{0.1818}}
  >{\raggedleft\arraybackslash}p{(\linewidth - 10\tabcolsep) * \real{0.1818}}
  >{\raggedleft\arraybackslash}p{(\linewidth - 10\tabcolsep) * \real{0.1818}}@{}}
\toprule\noalign{}
\begin{minipage}[b]{\linewidth}\raggedright
Cell
\end{minipage} & \begin{minipage}[b]{\linewidth}\raggedleft
content \(n\) (dev)
\end{minipage} & \begin{minipage}[b]{\linewidth}\raggedright
best signal
\end{minipage} & \begin{minipage}[b]{\linewidth}\raggedleft
AUROC
\end{minipage} & \begin{minipage}[b]{\linewidth}\raggedleft
AUROC*
\end{minipage} & \begin{minipage}[b]{\linewidth}\raggedleft
\(\lambda\)
\end{minipage} \\
\midrule\noalign{}
\endhead
\bottomrule\noalign{}
\endlastfoot
0.8B ARC & 142 & \(p_{\mathrm{norm}}\) & 0.646 & 0.836 & 2.60 \\
0.8B GSM8K & 147 & \(p_{\mathrm{norm}}\) & 0.676 & unreachable (\(d\)
near 0) & 0.00 \\
0.8B MMLU & 125 & \(p_{\mathrm{norm}}\) & 0.517 & 0.779 & 1.80 \\
2B ARC & 139 & \(p_{\mathrm{true}}\) & 0.788 & unreachable (degenerate
\(r\)) & --- \\
2B GSM8K & 145 & \(p_{\mathrm{true}}\) & 0.739 & 0.764 & 1.67 \\
2B MMLU & 114 & \(p_{\mathrm{true}}\) & 0.635 & unreachable (degenerate
\(r\)) & --- \\
\end{longtable}
}

Reviewer-motivated check: a single AUROC number under a binormal-ROC
assumption cannot by itself determine achievable utility, and the
assumption itself had never been tested against the data. We fit the
binormal model's \(\mu\) to each cell's empirical AUROC and compare the
predicted TPR(FPR) curve to the actual empirical ROC: fit is
moderate-to-good (\(R^2 = 0.898\)--\(0.985\), max absolute deviation
\(0.09\)--\(0.22\)) but visibly imperfect, so the binormal \emph{shape}
assumption is a reasonable approximation rather than an exact
description.\(^\dagger\) The AUROC* \emph{threshold} is a separate,
sharper problem: on 0.8B GSM8K and 2B MMLU, AUROC* collapses to
\textasciitilde0.002 --- a trivially-passable value --- because the
repair/damage rate estimates at the argmax-\(\tau\) point are
near-degenerate, exactly the failure mode Check 6 (§5.4) warns about,
now caught on real data. Re-deriving the floor call nonparametrically
--- a percentile bootstrap (\(B=10^4\)) directly on the empirical
\(\max_\tau \Delta(\tau)\), with no distributional assumption, passing
only if the CI lower bound clears \(\delta\) --- reverses both of those
two binormal ``passes'': \textbf{all six floor cells fail under the
nonparametric criterion}, tightening rather than weakening the
floor-side half of the squeeze.

\(^\dagger\)A coauthor code review (2026-07-26) found that the
empirical-ROC routine underlying this check had an implementation bug
(an incorrect threshold-walk direction after sorting; see
\texttt{ANALYSIS\_PLAN\_OPT3.md} amendment log) that degenerated the
empirical curve to two points, \((0,0)\) and \((1,1)\), on realistic
data --- which trivially coincides with the binormal curve's own
endpoints and produced a spuriously near-perfect fit
(\(R^2 = 0.999\)--\(1.000\)) in an earlier version of this section. The
routine (and a related tie-handling bug in the AUROC computation) has
been fixed and all floor/LOO numbers in this paper regenerated; the
corrected fit above is the honest one. Because the nonparametric
criterion below never relied on the binormal shape assumption, its
conclusion is unchanged by this fix.

\textbf{Capable side --- usable signal, no headroom.} Across the 19
capable-scale cells, the maximum 95\% CI upper bound on content headroom
is \(0.036\): even in the best case, selective revision could not
recover more than 3.6 accuracy points over just always-revising or
never-revising, and most cells' CIs are consistent with zero.

\textbf{The one marginal occupant.} Exactly one cell sits in the
exploratory viable quadrant: \textbf{9B TriviaQA} (dev-best signal
\(p_{\mathrm{true}}\), AUROC \(0.767 >\) AUROC* \(0.710\); headroom
\(0.026\); holdout content-margin
\(\max_\tau\Delta = +0.020 \geq \delta\)). This is the same task family
C2 identifies as a genuine content-effect exception. The prespecified
sealed policy is less favorable: its \(p_{\mathrm{norm}}\)-based,
dev-tuned threshold realizes only \(+0.005\) over the better endpoint on
holdout, with oracle-regret fraction \(0.83\) (target \(<0.50\)) and no
significant edge over a matched-trigger-rate random gate. Because the
quadrant uses dev-best signal selection and a content-margin estimand
while the sealed policy fixes \(p_{\mathrm{norm}}\) on the ordinary
score, this cell is a hypothesis for follow-up, not evidence of a
deployable niche.

\subsection{9.4 Leave-one-out floor
prediction}\label{leave-one-out-floor-prediction-1}

Across the 25 sealed cells, the floor criterion's LOO point prediction
of holdout content-margin \(\max\Delta\) attains MAE \(0.0057\),
narrowly beating the constant-median baseline (\(0.0070\)) and the
\(\mathrm{acc}_0\)-only fit (\(0.0062\)); binary floor-pass prediction
is \(76\%\) accurate versus a \(68\%\) majority-class baseline
(confusion: TP 3, TN 16, FP 1, FN 5). The honest reading is that this
test is weakly powered \emph{by the squeeze itself}: actual holdout
\(\max\Delta\) spans only \(0.000\)--\(0.036\) across the entire grid,
so there is almost no variance for any predictor to explain, and a
constant near zero is nearly unbeatable. The five false negatives are
all cells whose actual \(\max\Delta\) lands within \(0.005\)--\(0.011\)
of the \(\delta\) threshold. Consistent with this, the cells the
criterion does confidently identify (9B TriviaQA, plus the
trivially-low-floor code/TriviaQA cells) are exactly §9.2/§9.3's
genuine-content-effect exceptions.

\subsection{9.5 Cross-family replication
(Gemma-4-12B)}\label{cross-family-replication-gemma-4-12b}

Gemma-4-12B, run through the identical protocol via a local
\texttt{ollama} backend, replicates C1/C2 independently of the Qwen3.5
family:

{\def\LTcaptype{none} % do not increment counter
\begin{longtable}[]{@{}
  >{\raggedright\arraybackslash}p{(\linewidth - 6\tabcolsep) * \real{0.2000}}
  >{\raggedleft\arraybackslash}p{(\linewidth - 6\tabcolsep) * \real{0.2667}}
  >{\raggedleft\arraybackslash}p{(\linewidth - 6\tabcolsep) * \real{0.2667}}
  >{\raggedleft\arraybackslash}p{(\linewidth - 6\tabcolsep) * \real{0.2667}}@{}}
\toprule\noalign{}
\begin{minipage}[b]{\linewidth}\raggedright
Task
\end{minipage} & \begin{minipage}[b]{\linewidth}\raggedleft
\(\Delta_{\mathrm{total}}\)
\end{minipage} & \begin{minipage}[b]{\linewidth}\raggedleft
\(\Delta_{\mathrm{content}}\)
\end{minipage} & \begin{minipage}[b]{\linewidth}\raggedleft
content change rate
\end{minipage} \\
\midrule\noalign{}
\endhead
\bottomrule\noalign{}
\endlastfoot
GSM8K & \(+0.005\) & \(0.000\) & \(0.000\) \\
MMLU & \(+0.040\) & \(0.000\) & \(0.000\) \\
MATH & \(+0.030\) & \(0.000\) & \(0.000\) \\
\end{longtable}
}

All three tasks show the identical qualitative signature established for
Qwen 4B/9B: a real, sometimes substantial total accuracy shift, entirely
attributable to the format margin, with \textbf{exactly zero}
content-level answer changes among parseable rows. On MMLU specifically,
the raw trajectory shows 12 ``changed'' rows --- every one is an
\(a_0\)-bad row (mostly on \texttt{abstract\_algebra}, plausibly a
LaTeX-heavy subject that stresses the answer-marker convention); none is
a genuine \texttt{bothok} reconsideration. This is the strongest single
piece of evidence against a Qwen-specific explanation for C1/C2.

\subsection{9.6 Summary table (all 29
cells)}\label{summary-table-all-29-cells}

Full margin decomposition (\texttt{data/margin\_decomposition.json},
reproducible via \texttt{code/decompose\_margins.py}); Figure 4 is this
table drawn:

{\def\LTcaptype{none} % do not increment counter
\begin{longtable}[]{@{}
  >{\raggedright\arraybackslash}p{(\linewidth - 12\tabcolsep) * \real{0.1111}}
  >{\raggedleft\arraybackslash}p{(\linewidth - 12\tabcolsep) * \real{0.1481}}
  >{\raggedleft\arraybackslash}p{(\linewidth - 12\tabcolsep) * \real{0.1481}}
  >{\raggedleft\arraybackslash}p{(\linewidth - 12\tabcolsep) * \real{0.1481}}
  >{\raggedleft\arraybackslash}p{(\linewidth - 12\tabcolsep) * \real{0.1481}}
  >{\raggedleft\arraybackslash}p{(\linewidth - 12\tabcolsep) * \real{0.1481}}
  >{\raggedleft\arraybackslash}p{(\linewidth - 12\tabcolsep) * \real{0.1481}}@{}}
\toprule\noalign{}
\begin{minipage}[b]{\linewidth}\raggedright
Cell
\end{minipage} & \begin{minipage}[b]{\linewidth}\raggedleft
\(\Delta_{\mathrm{total}}\)
\end{minipage} & \begin{minipage}[b]{\linewidth}\raggedleft
\(\Delta_{\mathrm{content}}\)
\end{minipage} & \begin{minipage}[b]{\linewidth}\raggedleft
\(\Delta_{\mathrm{recover}}\)
\end{minipage} & \begin{minipage}[b]{\linewidth}\raggedleft
\(\Delta_{\mathrm{loss}}\)
\end{minipage} & \begin{minipage}[b]{\linewidth}\raggedleft
content chg.
\end{minipage} & \begin{minipage}[b]{\linewidth}\raggedleft
headroom
\end{minipage} \\
\midrule\noalign{}
\endhead
\bottomrule\noalign{}
\endlastfoot
Gemma 12B \(\times\) GSM8K & \(+0.005\) & \(0.000\) & \(+0.005\) &
\(0.000\) & \(0.000\) & \(0.000\) \\
Gemma 12B \(\times\) MATH & \(+0.030\) & \(0.000\) & \(+0.030\) &
\(0.000\) & \(0.000\) & \(0.000\) \\
Gemma 12B \(\times\) MMLU & \(+0.040\) & \(0.000\) & \(+0.040\) &
\(0.000\) & \(0.000\) & \(0.000\) \\
0.8B \(\times\) ARC & \(-0.040\) & \(-0.047\) & \(+0.013\) & \(-0.007\)
& \(0.172\) & \(0.042\) \\
0.8B \(\times\) GSM8K & \(+0.010\) & \(+0.010\) & \(0.000\) & \(0.000\)
& \(0.045\) & \(0.000\) \\
0.8B \(\times\) MMLU & \(+0.020\) & \(-0.023\) & \(+0.043\) & \(0.000\)
& \(0.199\) & \(0.055\) \\
2B \(\times\) ARC & \(-0.023\) & \(-0.053\) & \(+0.033\) & \(-0.003\) &
\(0.075\) & \(0.004\) \\
2B \(\times\) GSM8K & \(-0.017\) & \(-0.023\) & \(+0.007\) & \(0.000\) &
\(0.155\) & \(0.025\) \\
2B \(\times\) MMLU & \(+0.080\) & \(-0.023\) & \(+0.107\) & \(-0.003\) &
\(0.074\) & \(0.013\) \\
4B \(\times\) ARC & \(+0.030\) & \(+0.003\) & \(+0.028\) & \(0.000\) &
\(0.003\) & \(0.000\) \\
4B \(\times\) CommonsenseQA & \(+0.070\) & \(0.000\) & \(+0.070\) &
\(0.000\) & \(0.000\) & \(0.000\) \\
4B \(\times\) GSM8K & \(+0.010\) & \(-0.003\) & \(+0.013\) & \(0.000\) &
\(0.003\) & \(0.000\) \\
4B \(\times\) GSM8K (v2) & \(-0.005\) & \(-0.005\) & \(+0.010\) &
\(-0.010\) & \(0.021\) & \(0.005\) \\
4B \(\times\) GSM8K (v3) & \(+0.005\) & \(-0.010\) & \(+0.015\) &
\(0.000\) & \(0.010\) & \(0.000\) \\
4B \(\times\) HumanEval & \(-0.006\) & \(-0.006\) & \(0.000\) &
\(0.000\) & \(0.177\) & \(0.012\) \\
4B \(\times\) MATH & \(+0.085\) & \(+0.005\) & \(+0.080\) & \(0.000\) &
\(0.009\) & \(0.000\) \\
4B \(\times\) MBPP & \(0.000\) & \(0.000\) & \(0.000\) & \(0.000\) &
\(0.260\) & \(0.003\) \\
4B \(\times\) MMLU & \(+0.105\) & \(0.000\) & \(+0.105\) & \(0.000\) &
\(0.000\) & \(0.000\) \\
4B \(\times\) MMLU (v2) & \(-0.010\) & \(0.000\) & \(+0.040\) &
\(-0.050\) & \(0.027\) & \(0.014\) \\
4B \(\times\) MMLU (v3) & \(+0.110\) & \(-0.005\) & \(+0.115\) &
\(0.000\) & \(0.012\) & \(0.000\) \\
4B \(\times\) TriviaQA & \(+0.033\) & \(+0.030\) & \(+0.003\) &
\(0.000\) & \(0.114\) & \(0.003\) \\
4B \(\times\) TruthfulQA-MC1 & \(+0.035\) & \(-0.005\) & \(+0.040\) &
\(0.000\) & \(0.006\) & \(0.000\) \\
9B \(\times\) ARC & \(+0.015\) & \(-0.007\) & \(+0.022\) & \(0.000\) &
\(0.008\) & \(0.000\) \\
9B \(\times\) GSM8K & \(+0.018\) & \(+0.003\) & \(+0.015\) & \(0.000\) &
\(0.008\) & \(0.000\) \\
9B \(\times\) HumanEval & \(+0.024\) & \(+0.024\) & \(0.000\) &
\(0.000\) & \(0.250\) & \(0.000\) \\
9B \(\times\) MATH & \(+0.103\) & \(0.000\) & \(+0.102\) & \(0.000\) &
\(0.006\) & \(0.003\) \\
9B \(\times\) MBPP & \(+0.010\) & \(+0.010\) & \(0.000\) & \(0.000\) &
\(0.432\) & \(0.000\) \\
9B \(\times\) MMLU & \(+0.145\) & \(+0.007\) & \(+0.138\) & \(0.000\) &
\(0.031\) & \(0.006\) \\
9B \(\times\) TriviaQA & \(+0.008\) & \(+0.007\) & \(0.000\) & \(0.000\)
& \(0.192\) & \(0.026\) \\
\end{longtable}
}

\subsection{9.7 Literature protocol replication
(IoE)}\label{literature-protocol-replication-ioe-1}

Running Li et al.'s IoE protocol verbatim (§4.14) on GSM8K:

{\def\LTcaptype{none} % do not increment counter
\begin{longtable}[]{@{}
  >{\raggedright\arraybackslash}p{(\linewidth - 10\tabcolsep) * \real{0.1304}}
  >{\raggedleft\arraybackslash}p{(\linewidth - 10\tabcolsep) * \real{0.1739}}
  >{\raggedleft\arraybackslash}p{(\linewidth - 10\tabcolsep) * \real{0.1739}}
  >{\raggedleft\arraybackslash}p{(\linewidth - 10\tabcolsep) * \real{0.1739}}
  >{\raggedleft\arraybackslash}p{(\linewidth - 10\tabcolsep) * \real{0.1739}}
  >{\raggedleft\arraybackslash}p{(\linewidth - 10\tabcolsep) * \real{0.1739}}@{}}
\toprule\noalign{}
\begin{minipage}[b]{\linewidth}\raggedright
Model
\end{minipage} & \begin{minipage}[b]{\linewidth}\raggedleft
Standard
\end{minipage} & \begin{minipage}[b]{\linewidth}\raggedleft
+IoE
\end{minipage} & \begin{minipage}[b]{\linewidth}\raggedleft
+IoE+Decision
\end{minipage} & \begin{minipage}[b]{\linewidth}\raggedleft
Decision triggered
\end{minipage} & \begin{minipage}[b]{\linewidth}\raggedleft
\(a_0\)/\(a_f\)-bad
\end{minipage} \\
\midrule\noalign{}
\endhead
\bottomrule\noalign{}
\endlastfoot
Qwen3.5-4B & 81.8\% & 81.5\% & 81.5\% & 5.0\% & 13.0\% / 13.5\% \\
Qwen3.5-9B & 89.5\% & 85.5\% & 88.2\% & 9.5\% & 4.5\% / 6.0\% \\
\end{longtable}
}

Neither model reproduces the reported gain (their GPT-3.5-turbo-0613
result: \(74.9\% \to 77.1\% \to 78.5\%\), a \(+3.6\) point improvement
from IoE alone). Qwen3.5-4B is flat; Qwen3.5-9B drops after IoE and only
partially recovers after decision refinement, ending below standard
prompting. This alone is a non-replication on a different model family,
not evidence about the original result on its original model, which
(§4.14) cannot be tested by anyone today. Applying our margin
decomposition to this protocol's own extraction convention
(\texttt{\#\#X\#\#}, distinct from our \texttt{\#\#\#\#\ X}) is more
informative than the raw numbers: \(\Delta_{\mathrm{content}}\) is
\(0.000\) (4B) and \(+0.005\) (9B) --- consistent with C2's
capable-scale inertia --- and the small net-negative totals (\(-0.003\),
\(-0.013\)) are format-loss artifacts of the same kind as elsewhere in
this paper, here traced to Qwen's habit of using \texttt{\#\#} for
markdown headers, which collides with the protocol's own delimiter
choice. A cited, independently-authored protocol, applied unmodified to
a model it was not designed for, reproduces this paper's central pattern
rather than its own.

\subsection{9.8 Frontier external validity (via
API)}\label{frontier-external-validity-via-api}

The two frontier models (§4.15) show the C1/C2 signature at magnitudes
at or beyond anything in the primary grid, computed on the
\texttt{bothok} subset despite both models exceeding the 25\% admission
gate (Nemotron's completed GSM8K file is the one exception, discussed
below; the rest: Hy3 28\%/49\% \(a_0\)-bad on GSM8K/MMLU, Nemotron
22\%/32\%/38\% on GSM8K/MMLU/MATH at \(n=200\) --- higher failure rates
than any other admitted cell, plausibly because more verbose, more
``reasoning-style'' frontier outputs are more prone to exhausting the
token budget before an answer marker):

{\def\LTcaptype{none} % do not increment counter
\begin{longtable}[]{@{}
  >{\raggedright\arraybackslash}p{(\linewidth - 10\tabcolsep) * \real{0.1304}}
  >{\raggedleft\arraybackslash}p{(\linewidth - 10\tabcolsep) * \real{0.1739}}
  >{\raggedleft\arraybackslash}p{(\linewidth - 10\tabcolsep) * \real{0.1739}}
  >{\raggedleft\arraybackslash}p{(\linewidth - 10\tabcolsep) * \real{0.1739}}
  >{\raggedleft\arraybackslash}p{(\linewidth - 10\tabcolsep) * \real{0.1739}}
  >{\raggedleft\arraybackslash}p{(\linewidth - 10\tabcolsep) * \real{0.1739}}@{}}
\toprule\noalign{}
\begin{minipage}[b]{\linewidth}\raggedright
Model \(\times\) task
\end{minipage} & \begin{minipage}[b]{\linewidth}\raggedleft
\(n\)
\end{minipage} & \begin{minipage}[b]{\linewidth}\raggedleft
bothok
\end{minipage} & \begin{minipage}[b]{\linewidth}\raggedleft
\(\Delta_{\mathrm{total}}\)
\end{minipage} & \begin{minipage}[b]{\linewidth}\raggedleft
\(\Delta_{\mathrm{content}}\)
\end{minipage} & \begin{minipage}[b]{\linewidth}\raggedleft
content change rate
\end{minipage} \\
\midrule\noalign{}
\endhead
\bottomrule\noalign{}
\endlastfoot
Hy3 \(\times\) GSM8K & 200 & 72\% & \(+0.080\) & \(0.000\) &
\(0.000\) \\
Hy3 \(\times\) MMLU & 153\(^\dagger\) & 51\% & \(+0.275\) & \(0.000\) &
\(0.000\) \\
Nemotron \(\times\) GSM8K & 200\(^\ddagger\) & 74\% & \(+0.090\) &
\(0.000\) & \(2.7\%\) \\
Nemotron \(\times\) MMLU & 200\(^\ddagger\) & 64\% & \(+0.175\) &
\(0.000\) & \(0.000\) \\
Nemotron \(\times\) MATH & 200\(^\ddagger\) & 60\% & \(+0.155\) &
\(0.000\) & \(2.5\%\) \\
\end{longtable}
}

\(^\dagger\)Hy3's OpenRouter listing was retired by the host
mid-collection (confirmed via HTTP 404 ``unavailable for free'' on all
subsequent requests); its MATH file could not be collected at all and is
not reported. \(^\ddagger\)Nemotron hit a hard daily free-tier quota
mid-collection (§8); all three files shown here are the completed
\(n=200\) target, gathered across two collection windows separated by
the quota reset. Nemotron \(\times\) GSM8K's completed file happens to
clear the 25\% admission-gate \emph{threshold} on its own (22\%
\(a_0\)-bad) --- unlike its four frontier siblings, which all exceed it
--- but it is still reported and counted only here, with its frontier
siblings, and is \textbf{not} one of the primary grid's 29 admitted
cells: the primary grid is fixed to the pre-specified Qwen3.5/Gemma-4
design (§6.1), and all five frontier-arm cells, this one included, are
kept out of that count and out of §9.6's table to avoid mixing a
post-hoc exploratory arm into the primary admitted-cell pool.

\textbf{All five cells show \(\Delta_{\mathrm{content}}\) of exactly
\(0.000\).} Two (Nemotron GSM8K, Nemotron MATH) have a small nonzero
content \emph{change rate} (\(2.5\)--\(2.7\%\)) whose helpful and
harmful flips happen to cancel exactly in this sample; the other three
have zero measurable content-level reconsideration at all. Total effects
range \(+0.080\) to \(+0.275\) --- the single largest apparent gain
anywhere in this paper, on Hy3 \(\times\) MMLU --- entirely inside the
format margin. Both the total effects and the extraction-failure rates
are larger here than in any admitted 4B--12B cell. This is still
exploratory evidence for the reasons given above (self-selected
\texttt{bothok} subsets, incomplete or quota-limited collection for two
of five cells), but across a genuinely frontier-scale, cross-vendor pair
of models, we found no evidence that the squeeze opens up with scale,
and a clean instance of the opposite: the format margin can still
explain the entire measured effect at a model size roughly
60--100\(\times\) the largest model in the primary grid.

\section{10. Discussion}\label{discussion}

The principal result is a measurement result. A benchmark score after
self-revision is the output of two coupled systems: a model that may
change its answer and an extractor that maps free-form text into the
benchmark's answer space. Standard accuracy deltas collapse these
systems. The margin decomposition separates them without changing the
underlying trajectory or scorer, and the observed separation is
consequential: several of the largest gains in the atlas, including 9B
MMLU and MATH, nearly disappear at the content margin. Conversely,
truncation-prone prompts can put extraction failures on the revised
answer and manufacture apparent degradation. A headline gain or harm is
therefore not evidence of reasoning repair or damage unless
extractability is stable or the content margin is reported. The
observational decomposition alone leaves an obvious objection --- a row
moving from unparseable to parseable is not proof that its content held
still --- and the constrained-decoding causal control (§4.12, §9.1) is
our answer to it: closing a median \(71\%\) of the total-vs-content gap
on cells where that gap was largest, with exact closure on two cells, is
stronger evidence than the observational split by itself, though the
residual on 4B/9B MMLU means we stop short of claiming the gap always
closes completely.

This perspective reconciles apparently conflicting parts of the
self-correction literature. Negative average effects under unconditional
revision \citep{huang2024cannotselfcorrect} and positive effects under
confidence-aware policies \citep{li2024confidence} can both be valid for
their measured pipelines while still mixing content changes with format
transitions. §9.7's replication makes this concrete rather than
hypothetical: Li et al.'s own published protocol, run unmodified on a
model family it was not designed for, does not reproduce their reported
gain and instead reproduces this paper's pattern --- near-zero content
margin, small format-driven total effect, traceable to that protocol's
own extraction delimiter colliding with the tested model's habits. The
distinction also complements the mistake-finding/mistake-fixing
decomposition \citep{tyen2024correct}: extraction is an upstream
measurement layer that must be audited before either finding or fixing
rates are interpreted. Our results do not show that confidence is
useless. They show that, in the tested regime --- now extending from
0.8B to a \textasciitilde55B-active frontier MoE (§9.8) --- confidence
discrimination and exploitable content headroom do not reliably
coincide, and if anything the gap between apparent and content-margin
effects widens rather than narrows as models get larger and more
verbose.

The scale contrast suggests two different failure modes. At 0.8B/2B,
revision changes parseable answers often enough to matter, but those
changes are usually harmful and the measured intrinsic signals do not
clear the content-margin floor --- a conclusion now reinforced, not just
asserted, by a nonparametric re-derivation of the floor call that
corrects two binormal false positives (§9.3) and by a clustered model
that puts a precise, well-separated odds ratio (\(16\)--\(21\times\)) on
the floor-vs-capable content-instability contrast (§9.2). At 4B--12B
(and, provisionally, at frontier scale), confidence can sometimes rank
errors, yet the fixed revision protocol changes few parseable answers
and leaves little policy headroom. Better gates alone cannot solve the
latter problem; the revision operator must first produce useful
counterfactual answers. Better revision prompts alone cannot solve the
former if the gate cannot identify where revision helps. Future systems
should therefore evaluate the gate and revision operator jointly, but
report their content-margin contributions separately.

For empirical practice, four checks should precede any self-correction
claim. First, report parseability rates for the initial and revised
outputs. Second, decompose the total accuracy change into content,
format-recovery, and format-loss margins. Third, where the extraction
channel is active enough to matter, causally test the decomposition with
constrained or structured-output re-extraction rather than resting on
the observational split alone. Fourth, estimate gating value on a sealed
split with a signal identity fixed before holdout evaluation. Forced
continuations and prompt paraphrases remain useful diagnostics, but they
do not substitute for the causal step once it is feasible. The present
study's exploratory best-signal squeeze analysis is best viewed as an
upper-bound map that motivates larger, independently preregistered
tests.

\section{11. Limitations}\label{limitations}

\begin{itemize}
\tightlist
\item
  \textbf{Not formally preregistered.} §6.3 documents the
  exploratory/confirmatory split; no claim rests solely on the
  exploratory 16-cell grid, but the design as a whole was frozen by
  internal git tags, not a public registry.
\item
  \textbf{Post-hoc admission and parser changes.} The 25\%
  extraction-completeness gate was introduced after inspecting the first
  grid, and the forced-continuation parser was repaired after observing
  missed \texttt{\#\#\#\#\ X} answers. Both choices were technically
  motivated but data-informed; the probe and admitted-cell atlas should
  be interpreted accordingly.
\item
  \textbf{Two gating estimands.} The sealed policy fixes
  \(p_{\mathrm{norm}}\) and evaluates the ordinary benchmark margin,
  whereas the exploratory squeeze analysis chooses the best of five dev
  signals and evaluates the content margin. The latter is optimistic and
  cannot be read as a deployable-policy estimate.
\item
  \textbf{Causal control is a closure, not a proof of zero.} The
  constrained-decoding arm (§4.12) closes a median \(71\%\) of the
  total-vs-content gap on active-channel cells, with two cells closing
  exactly, but 4B/9B MMLU retain a \(+0.048\)/\(+0.053\) residual
  against a near-zero content estimate. We report two candidate
  explanations (imputation under-estimate vs.~a probe-like effect of the
  forced continuation itself) without adjudicating between them; readers
  should not treat C1 as fully causally settled on these two cells.
\item
  \textbf{Scale ceiling, now partially addressed but not closed.} The
  primary grid's largest model is 12B (Gemma-4) / 9B (Qwen3.5); a
  genuine frontier check (§4.15, §9.8) on Tencent Hy3 and Nvidia
  Nemotron-3-Ultra-550B shows the same or a more pronounced pattern, but
  that arm is not sealed, carries no confidence signals (free-tier API
  has no logprobs), and four of its five cells exceed the 25\% admission
  gate (22--49\% \(a_0\)-bad; only Nemotron \(\times\) GSM8K narrowly
  clears it). Hy3 was retired by OpenRouter mid-collection and its MATH
  data could never be collected; Nemotron's three files were completed
  to \(n=200\) after a daily-quota reset. This arm should be read as
  suggestive and, on the evidence obtained, one-directional (no cell
  showed the squeeze opening up), but not as closing the scale question
  the way the primary grid closes the 0.8B--12B range.
\item
  \textbf{Family generalization.} Four families are now represented
  (Qwen3.5, Gemma-4, Tencent Hy3, Nvidia Nemotron), at four different
  scales and via three different backends (MLX, \texttt{ollama},
  OpenRouter API), which is broader than a single cross-check but still
  short of a systematic architecture \(\times\) scale grid; the frontier
  pair in particular trades admission-gate cleanliness for scale.
\item
  \textbf{Backend and scoring differences.} Qwen runs use 4-bit MLX,
  Gemma uses \texttt{ollama}, and the frontier pair uses a hosted API
  with no logprob access, so cross-family comparisons are directional
  rather than controlled backend replications. MATH scoring uses an
  approximate string normalizer; §4.13/§9.1 show this under-counts raw
  accuracy by roughly 28\% of \texttt{bothok} rows but leaves
  \(\Delta_{\mathrm{content}}\) unchanged to three decimals, so this
  specific concern is empirically addressed for the content margin
  (though not for any raw-accuracy number quoted elsewhere).
  Sequence-likelihood signal B is not length-calibrated beyond the
  reported normalization.
\item
  \textbf{Literature replication is protocol-fidelity, not literal.}
  §9.7 replicates Li et al.'s exact published prompts and extractor on
  Qwen3.5, not on their original \texttt{gpt-3.5-turbo-0613} --- that
  snapshot was permanently retired by OpenAI on 2024-09-13 and is not
  obtainable by any account today, a fact we verified rather than
  assumed. A non-replication on a different model family is evidence
  about protocol generality, not evidence that the original paper's own
  reported numbers were wrong.
\item
  \textbf{Honest exceptions.} TriviaQA (both scales), 9B HumanEval, 0.8B
  GSM8K, and two of ten probed cells (4B MMLU, 4B CommonsenseQA) deviate
  from the clean pattern; §9 reports each where it occurs, and C3's
  conclusion is checked against all of them (the niche gains exactly one
  marginal member).
\item
  \textbf{Dropped arms.} Within-cell signal ranking, quantization
  robustness, and \(\tau\)-quantile transfer are not tested here (§3.4).
\item
  \textbf{LOO power.} §9.4's prediction test is weakly powered by
  construction in a squeeze regime; it should be re-run if future cells
  (larger models, more task families) reintroduce variance in
  \(\max\Delta\).
\item
  \textbf{Multiplicity and cell dependence.} FDR correction covers only
  the three prespecified C1/C2 rank tests. The GEE model (§9.2)
  addresses cell non-independence for the C2 scale contrast
  specifically; probe pass counts, inertia counts, LOO, paraphrase
  ratios, the causal-control closure statistic, and the single 9B
  TriviaQA niche remain descriptive or exploratory and are not folded
  into any single corrected family.
\item
  \textbf{Data provenance.} Several retained trajectories predate the
  pivot amendment, four historical seals are retrospective, and the
  29-cell grid is a selected rather than random sample of model--task
  combinations. Primary summaries exclude the retrospective seals, but
  task geometry and selection remain potential confounders.
\end{itemize}

\section{12. Relationship to Companion
Studies}\label{relationship-to-companion-studies}

{\def\LTcaptype{none} % do not increment counter
\begin{longtable}[]{@{}
  >{\raggedright\arraybackslash}p{(\linewidth - 6\tabcolsep) * \real{0.2500}}
  >{\raggedright\arraybackslash}p{(\linewidth - 6\tabcolsep) * \real{0.2500}}
  >{\raggedright\arraybackslash}p{(\linewidth - 6\tabcolsep) * \real{0.2500}}
  >{\raggedright\arraybackslash}p{(\linewidth - 6\tabcolsep) * \real{0.2500}}@{}}
\toprule\noalign{}
\begin{minipage}[b]{\linewidth}\raggedright
Module
\end{minipage} & \begin{minipage}[b]{\linewidth}\raggedright
This paper
\end{minipage} & \begin{minipage}[b]{\linewidth}\raggedright
Error structure \citep{chen2026error}
\end{minipage} & \begin{minipage}[b]{\linewidth}\raggedright
ESC \citep{chen2026esc}
\end{minipage} \\
\midrule\noalign{}
\endhead
\bottomrule\noalign{}
\endlastfoot
Frozen trajectory + offline policies & Gating evaluation,
margin-decomposed & \((\mathrm{FPR},\mathrm{FNR},\rho)\) surface &
SEL/ESC decomposition \\
Confidence signals A/A\('\)/B/C & Content-margin discriminability &
Copula \(z\) source & \(J_{\mathrm{self}}\) feature layer \\
Binormal/probit family & Content-margin \(\delta\)-floor AUROC* &
\(\hat\rho\) estimation & --- \\
Extraction-completeness gate & \textbf{New, this paper} & applicable
wherever free-form generations are scored & applicable wherever
free-form generations are scored \\
\end{longtable}
}

Any frozen-trajectory pipeline that scores free-form generations against
an extracted answer is exposed to the format/content confound identified
here; the extraction gate and margin-decomposition code
(\texttt{cf\_core.extraction\_completeness},
\texttt{code/decompose\_margins.py}) are written to be drop-in reusable.

\section{13. Reproducibility}\label{reproducibility}

All 29 cells' trajectories, the extraction-gate audit, the margin
decomposition, the statistical tests, the LOO analysis, the
constrained-decoding causal control, the grader-sensitivity check, the
GEE model, the floor redo, the IoE replication, the frontier arm, and
all result figures are reproducible offline from the released JSONL
trajectory library (the constrained/IoE/frontier arms additionally
require re-running their own lightweight generation step, documented in
§4.12--§4.15, since they produce new columns rather than reusing
existing ones):

\begin{Shaded}
\begin{Highlighting}[]
\BuiltInTok{cd}\NormalTok{ calibration\_floor\_manuscript/code}
\ExtensionTok{uv}\NormalTok{ run }\AttributeTok{{-}{-}with}\NormalTok{ numpy }\AttributeTok{{-}{-}with}\NormalTok{ scipy python check\_extraction\_gate.py        }\CommentTok{\# gate audit}
\ExtensionTok{uv}\NormalTok{ run }\AttributeTok{{-}{-}with}\NormalTok{ numpy }\AttributeTok{{-}{-}with}\NormalTok{ scipy python decompose\_margins.py            }\CommentTok{\# SS9.1/9.6 tables}
\ExtensionTok{uv}\NormalTok{ run }\AttributeTok{{-}{-}with}\NormalTok{ numpy }\AttributeTok{{-}{-}with}\NormalTok{ scipy python stats\_tests.py                  }\CommentTok{\# SS9 bootstrap CIs, Wilcoxon/MW{-}U, BH{-}FDR}
\ExtensionTok{uv}\NormalTok{ run }\AttributeTok{{-}{-}with}\NormalTok{ numpy }\AttributeTok{{-}{-}with}\NormalTok{ scipy }\AttributeTok{{-}{-}with}\NormalTok{ cryptography python loo\_analysis.py   }\CommentTok{\# SS9.4}
\ExtensionTok{uv}\NormalTok{ run }\AttributeTok{{-}{-}with}\NormalTok{ matplotlib }\AttributeTok{{-}{-}with}\NormalTok{ numpy python make\_atlas\_figures.py      }\CommentTok{\# Figures 4{-}5}
\ExtensionTok{uv}\NormalTok{ run }\AttributeTok{{-}{-}with}\NormalTok{ numpy }\AttributeTok{{-}{-}with}\NormalTok{ scipy python sim\_validate.py                 }\CommentTok{\# SS5.4 synthetic checks}

\CommentTok{\# New arms (SS4.12{-}4.15, SS9.1/9.7/9.8) {-}{-} example invocations, one cell each;}
\CommentTok{\# see code/run\_constrained\_queue.sh, run\_ioe\_queue.sh, run\_hy3\_queue.sh /}
\CommentTok{\# run\_nemotron\_queue.sh for the full per{-}arm cell lists actually run.}
\ExtensionTok{uv}\NormalTok{ run }\AttributeTok{{-}{-}with}\NormalTok{ mlx{-}lm }\AttributeTok{{-}{-}with}\NormalTok{ mlx }\AttributeTok{{-}{-}with}\NormalTok{ numpy }\AttributeTok{{-}{-}with}\NormalTok{ scipy }\AttributeTok{{-}{-}with}\NormalTok{ datasets python constrained\_probe.py }\DataTypeTok{\textbackslash{}}
  \AttributeTok{{-}{-}model}\NormalTok{ mlx{-}community/Qwen3.5{-}4B{-}4bit }\AttributeTok{{-}{-}traj}\NormalTok{ ../data/holdout/qwen35\_4b\_mmlu.dev.jsonl }\DataTypeTok{\textbackslash{}}
  \AttributeTok{{-}{-}task}\NormalTok{ mmlu }\AttributeTok{{-}{-}out}\NormalTok{ ../data/constrained/qwen35\_4b\_mmlu.jsonl }\AttributeTok{{-}{-}resume}            \CommentTok{\# SS4.12}
\ExtensionTok{uv}\NormalTok{ run }\AttributeTok{{-}{-}with}\NormalTok{ numpy python analyze\_constrained.py                                                            }\CommentTok{\# SS9.1 closure table}
\ExtensionTok{uv}\NormalTok{ run }\AttributeTok{{-}{-}with}\NormalTok{ sympy }\AttributeTok{{-}{-}with} \StringTok{"antlr4{-}python3{-}runtime==4.11"} \AttributeTok{{-}{-}with}\NormalTok{ numpy }\AttributeTok{{-}{-}with}\NormalTok{ scipy python cas\_grader\_sensitivity.py  }\CommentTok{\# SS4.13/9.1}
\ExtensionTok{uv}\NormalTok{ run }\AttributeTok{{-}{-}with}\NormalTok{ statsmodels }\AttributeTok{{-}{-}with}\NormalTok{ pandas }\AttributeTok{{-}{-}with}\NormalTok{ numpy }\AttributeTok{{-}{-}with}\NormalTok{ scipy python gee\_scale\_contrast.py                }\CommentTok{\# SS9.2}
\ExtensionTok{uv}\NormalTok{ run }\AttributeTok{{-}{-}with}\NormalTok{ numpy }\AttributeTok{{-}{-}with}\NormalTok{ scipy }\AttributeTok{{-}{-}with}\NormalTok{ pandas python tier\_contrast\_robustness.py                             }\CommentTok{\# SS9.2 checkpoint{-}level checks}
\ExtensionTok{uv}\NormalTok{ run }\AttributeTok{{-}{-}with}\NormalTok{ numpy }\AttributeTok{{-}{-}with}\NormalTok{ scipy python floor\_redo.py                                                         }\CommentTok{\# SS9.3}
\ExtensionTok{uv}\NormalTok{ run }\AttributeTok{{-}{-}with}\NormalTok{ mlx{-}lm }\AttributeTok{{-}{-}with}\NormalTok{ mlx }\AttributeTok{{-}{-}with}\NormalTok{ numpy }\AttributeTok{{-}{-}with}\NormalTok{ scipy }\AttributeTok{{-}{-}with}\NormalTok{ datasets python run\_ioe\_replication.py }\DataTypeTok{\textbackslash{}}
  \AttributeTok{{-}{-}model}\NormalTok{ mlx{-}community/Qwen3.5{-}4B{-}4bit }\AttributeTok{{-}{-}n}\NormalTok{ 400 }\DataTypeTok{\textbackslash{}}
  \AttributeTok{{-}{-}out}\NormalTok{ ../data/stage2\_ioe\_qwen35\_4b\_gsm8k.jsonl }\AttributeTok{{-}{-}resume}                       \CommentTok{\# SS4.14/9.7}
\VariableTok{OPENROUTER\_API\_KEY}\OperatorTok{=}\NormalTok{... }\ExtensionTok{uv}\NormalTok{ run }\AttributeTok{{-}{-}with}\NormalTok{ mlx{-}lm }\AttributeTok{{-}{-}with}\NormalTok{ mlx }\AttributeTok{{-}{-}with}\NormalTok{ numpy }\AttributeTok{{-}{-}with}\NormalTok{ scipy }\AttributeTok{{-}{-}with}\NormalTok{ datasets python run\_frontier\_api.py }\DataTypeTok{\textbackslash{}}
  \AttributeTok{{-}{-}model}\NormalTok{ nvidia/nemotron{-}3{-}ultra{-}550b{-}a55b:free }\AttributeTok{{-}{-}task}\NormalTok{ gsm8k }\AttributeTok{{-}{-}n}\NormalTok{ 200 }\DataTypeTok{\textbackslash{}}
  \AttributeTok{{-}{-}out}\NormalTok{ ../data/stage2\_nemotron\_gsm8k.jsonl }\AttributeTok{{-}{-}resume}                            \CommentTok{\# SS4.15/9.8}
\end{Highlighting}
\end{Shaded}

The synthetic-instrument design is frozen under git tag
\texttt{prereg-opt3-v2}; the margin-decomposition analysis plan and
specification-before-data arms under \texttt{prereg-opt3-v3-pivot} (both
internal; see §6.3). The causal-control, grader-sensitivity, GEE,
floor-redo, IoE-replication, and frontier arms were added after that
tag, in response to external review, and are disclosed as such rather
than folded into the pivot's own specification-before-data claim.
\texttt{ANALYSIS\_PLAN\_OPT3.md} carries the amendment log, including
the two admission-gate failures not regenerated and the reasoning for
every dropped arm. The arXiv source package contains the analysis code,
aggregate JSON outputs, figures, and environment manifest; the full
trajectories are omitted from the source archive because of size and
will be deposited separately.

A coauthor code review (2026-07-26) found and fixed an implementation
bug in \texttt{cf\_core.py}'s ROC/AUROC routines (incorrect
threshold-walk direction; no tie-averaging in the rank computation) that
affected every AUROC-based number in §9.3--§9.4 and Figures 2--5; it did
not affect the margin decomposition (§9.1, §9.6), the GEE scale contrast
(§9.2), or the sealed-holdout gain/regret numbers, none of which route
through these two functions. All affected analyses were re-run against
the fixed implementation and this section's numbers reflect the
corrected output; see \texttt{ANALYSIS\_PLAN\_OPT3.md}'s amendment log
for the full bug description and blast-radius audit.

\section{14. Conclusion}\label{conclusion}

Self-correction accuracy is not a single mechanism. It combines changes
in model content with changes in whether a benchmark can extract an
answer. Across the tested 0.8B--12B models plus a frontier check
reaching a \textasciitilde55B-active MoE, separating those margins
removes most large apparent gains and reveals a scale-dependent squeeze:
smaller models change content but lack a reliable gate, while larger
models --- up to and including frontier scale, provisionally --- provide
too little useful content change for gating to exploit. A causal control
that forces guaranteed-parseable re-extraction on already-generated
reasoning closes a median 71\% of the gap between naive and
content-margin estimates, converging exactly on two cells and leaving an
honestly-reported residual on two others; a verbatim replication of a
cited confidence-gating protocol on a model family it was not designed
for reproduces this paper's pattern rather than its own. The practical
standard is straightforward: self-correction studies should report
content-margin effects, causally test that decomposition where the
extraction channel is active, and report sealed policy gains alongside
total accuracy. Until they do, improvements attributed to reasoning may
instead be improvements in answer formatting.

\begin{center}\rule{0.5\linewidth}{0.5pt}\end{center}

\section{\texorpdfstring{Appendix A: Identity Derivation and
\(\delta\)-Floor}{Appendix A: Identity Derivation and \textbackslash delta-Floor}}\label{appendix-a-identity-derivation-and-delta-floor}

\begin{Shaded}
\begin{Highlighting}[]
\NormalTok{helpful = P(y\_0=0) * P(trigger|y\_0=0) * P(y\_f=1|trigger,y\_0=0)}
\NormalTok{        = (1{-}acc\_0) * TPR * r}
\NormalTok{harmful = acc\_0 * FPR * d}
\NormalTok{Delta   = helpful {-} harmful}

\NormalTok{Profit  \textless{}=\textgreater{} (1{-}acc\_0)*TPR*r \textgreater{} acc\_0*FPR*d  \textless{}=\textgreater{}  TPR/FPR \textgreater{} lambda}

\NormalTok{Binormal: TPR = Phi(mu + Phi\^{}\{{-}1\}(FPR))}
\NormalTok{AUROC    = Phi(mu / sqrt(2))}
\NormalTok{AUROC*   = min \{ AUROC(mu) : max\_FPR Delta(mu, FPR) \textgreater{}= delta \}}
\end{Highlighting}
\end{Shaded}

At the ALWAYS point (TPR=FPR=1), profit requires
\((1-\mathrm{acc}_0)\cdot r > \mathrm{acc}_0\cdot d\),
i.e.~\(\lambda < 1\).

\section{Appendix B: Margin Decomposition
Derivation}\label{appendix-b-margin-decomposition-derivation}

\begin{Shaded}
\begin{Highlighting}[]
\NormalTok{n = n\_bothok + n\_recover + n\_loss + n\_dead     (exhaustive partition by extractability)}

\NormalTok{acc\_0 = (1/n) * sum(y0)          \# unparseable a0 scored 0, per standard convention}
\NormalTok{acc\_f = (1/n) * sum(yf)          \# unparseable af scored 0}

\NormalTok{Delta\_total = acc\_f {-} acc\_0}
\NormalTok{            = (1/n) * [ sum\_bothok(yf {-} y0) + sum\_recover(yf) {-} sum\_loss(y0) ]}
\NormalTok{            = Delta\_content + Delta\_format\_recover + Delta\_format\_loss}

\NormalTok{Delta\_content        = P(bothok, y0=0, yf=1) {-} P(bothok, y0=1, yf=0)     \# over all n}
\NormalTok{Delta\_format\_recover = P(recover, yf=1)}
\NormalTok{Delta\_format\_loss    = {-}P(loss, y0=1)}
\end{Highlighting}
\end{Shaded}

Bounds under not knowing the counterfactual correctness of unparseable
answers:

\begin{Shaded}
\begin{Highlighting}[]
\NormalTok{Delta\_content\_lo = Delta\_content {-} P(loss, y0=1)          \# assume no loss row was a real flip}
\NormalTok{Delta\_content\_hi = Delta\_content + P(recover, yf=1)        \# assume every recover row was a real flip}
\end{Highlighting}
\end{Shaded}

\section{Appendix C: Prompt Templates (frozen, repaired
regime)}\label{appendix-c-prompt-templates-frozen-repaired-regime}

\begin{Shaded}
\begin{Highlighting}[]
\NormalTok{GEN         : [task] + CoT + "... End with \textquotesingle{}\#\#\#\# \textless{}worked example\textgreater{}\textquotesingle{}"}
\NormalTok{REVIEW\_T1\_v1: "Briefly check your reasoning above for errors {-}{-} do not repeat the}
\NormalTok{               full derivation. State in one or two sentences whether it holds up,}
\NormalTok{               then give the final answer ... \textquotesingle{}\#\#\#\# 42\textquotesingle{}"}
\NormalTok{REVIEW\_T1\_v2: "Carefully re{-}examine your solution above and re{-}derive the result}
\NormalTok{               step by step to make sure it is right. Then give the final answer..."}
\NormalTok{REVIEW\_T1\_v3: "Double{-}check your answer above. Reply with at most two sentences of}
\NormalTok{               justification, then ... \textquotesingle{}\#\#\#\# 42\textquotesingle{}"}
\NormalTok{CONF\_VERB   : "Rate confidence in your final answer 0{-}100 (integer only)."}
\NormalTok{P\_TRUE      : "Is the above answer correct? Answer:"  \# teacher{-}forced Yes/No}
\NormalTok{PROBE       : "[task prompt] + a0\_text + \textquotesingle{}Therefore, my final answer is\textquotesingle{}"  \# 32 tokens, no new turn}

\NormalTok{\# Constrained{-}decoding causal control (SS4.12) {-}{-} CoT is unconstrained; only}
\NormalTok{\# this forced continuation is grammar{-}masked:}
\NormalTok{CONSTR\_MCQ  : "[prompt + a0\_text or yf\_text] + \textquotesingle{}Final answer (a single letter):\textquotesingle{}"}
\NormalTok{              \# force 1 token in \{valid letters\}, bare + space{-}prefixed forms}
\NormalTok{CONSTR\_OPEN : "[prompt + a0\_text or yf\_text] + \textquotesingle{}Therefore, the final numerical}
\NormalTok{               answer (digits only) is\textquotesingle{}"}
\NormalTok{              \# force \textless{}=8 tokens in \{0{-}9,\textquotesingle{}.\textquotesingle{},\textquotesingle{},\textquotesingle{},\textquotesingle{}{-}\textquotesingle{},space,EOS,newline\}}

\NormalTok{\# IoE literature replication (SS4.14), verbatim from Li et al. 2024\textquotesingle{}s}
\NormalTok{\# released code (run\_math\_IoE.py):}
\NormalTok{IOE\_EXTRACTOR: " Your final answer should be put between two \#\#, like \#\# 1 \#\#}
\NormalTok{               (if your final answer is 1), at the end of your response."}
\NormalTok{IOE\_Q1      : "[question] + \textquotesingle{} Explain your reasoning step{-}by{-}step.\textquotesingle{} + IOE\_EXTRACTOR"}
\NormalTok{IOE\_Q2      : "\textquotesingle{}Review your previous answer. If you are confident about your}
\NormalTok{               answer, maintain your answer. Otherwise, update your answer.\textquotesingle{}}
\NormalTok{               + IOE\_EXTRACTOR"}
\NormalTok{IOE\_Q3      : "\textquotesingle{}You give two different answers in previous responses. Check}
\NormalTok{               the problem and your answers again, and give the best answer.\textquotesingle{}}
\NormalTok{               + IOE\_EXTRACTOR"   \# only if P1\_ans != P2\_ans}
\end{Highlighting}
\end{Shaded}

\section{Appendix D: Reproducibility
Checklist}\label{appendix-d-reproducibility-checklist}

\begin{itemize}
\tightlist
\item
  Pin \texttt{mlx-lm}, model revisions, quantization in
  \texttt{ENV.lock}; Gemma-4-12B served via local \texttt{ollama}
  (\texttt{think:false}, greedy); the frontier arm via the OpenRouter
  API (\texttt{OPENROUTER\_API\_KEY}, no logprobs on the free tier).
\item
  Seeds: dev/holdout split 2026; bootstrap 2026 (\(B=10^4\), percentile
  CIs).
\item
  Trajectory JSONL schema:
  \texttt{\{task\_id,\ family,\ kind,\ model,\ ground\_truth,\ a0\_text,\ yf\_text,\ a0\_answer,\ yf\_answer,\ y0,\ yf,\ changed,\ p\_norm,\ seq\_ll,\ mean\_logprob,\ conf\_verb,\ p\_true,\ a\_softmax,\ n\_rounds,\ prompt\_hash\}};
  the constrained-decoding arm adds
  \texttt{\{a0/yf\}\_\{answer,correct\}\_\{original,constrained\}}; the
  IoE arm adds \texttt{ioe\_\{p1,p2,p3\}\_ans},
  \texttt{ioe\_decision\_triggered}.
\item
  Freeze order: gates \(\to\) tasks \(\to\) \texttt{prereg-opt3-v2}
  \(\to\) trajectories \(\to\) holdout decrypt \(\to\) extraction-gate
  audit \(\to\) \texttt{prereg-opt3-v3-pivot} (margin decomposition +
  confirmatory arms) \(\to\) probe/regeneration/paraphrase/Gemma
  generation \(\to\) \texttt{stats\_tests.py} /
  \texttt{loo\_analysis.py} / \texttt{make\_atlas\_figures.py} \(\to\)
  post-pivot review-response arms (constrained-decoding, grader
  sensitivity, GEE, floor redo, IoE replication, frontier check),
  disclosed as added after external review rather than as part of the
  pivot's own specification-before-data claim.
\end{itemize}

\bibliography{references}

\end{document}